\documentclass[journal]{IEEEtran}

\usepackage{times}
\usepackage{graphicx}

\usepackage{amsmath,amssymb}
\usepackage{subfigure}
\usepackage{algorithm}
\usepackage{algorithmicx}
\usepackage{algpseudocode}
\usepackage{graphics}
\usepackage{threeparttable}
\usepackage{color}
\usepackage[normalem]{ulem}
\usepackage{multirow}
\usepackage{float}
\usepackage{amsfonts}
\usepackage{bm}
\usepackage{array}
\usepackage[table]{xcolor}
\usepackage{colortbl}
\usepackage{pifont}
\usepackage{diagbox}
\usepackage{rotating}
\usepackage{booktabs}
\usepackage{overpic}
\usepackage{textcomp}
\usepackage{contour}
\usepackage{url}
\usepackage{enumitem}
\usepackage{tabularx}
\usepackage{graphicx,epstopdf}
\usepackage[misc]{ifsym}

\usepackage{silence}
\def\eg{\emph{e.g.}}
\def\etc{\emph{etc}}
\def\etal{{\em et al.~}}

\usepackage{mathptmx}      
\DeclareMathAlphabet{\mathcal}{OMS}{cmsy}{m}{n}  
\DeclareSymbolFont{largesymbols}{OMX}{cmex}{m}{n}

\usepackage[pagebackref=false,breaklinks=true,colorlinks, bookmarks=false]{hyperref}

\usepackage{cleveref}
\crefformat{section}{\S#2#1#3}
\crefformat{subsection}{\S#2#1#3}
\crefformat{subsubsection}{\S#2#1#3}

\newcommand{\trb}[1]{\textbf{\textcolor{red}{#1}}}

\newcommand{\tbb}[1]{\textcolor{blue}{#1}}

\newcommand{\secref}[1]{Section \ref{#1}}
\newcommand{\tabref}[1]{Table \ref{#1}}
\newcommand{\figref}[1]{Figure \ref{#1}}
\newcommand{\equref}[1]{Equation \ref{#1}}

\newcommand{\ACSD}{ACSD~\cite{ju2014depth}}

\newcommand{\PCF}{PCF~\cite{chen2018progressively}}
\newcommand{\TANet}{TANet~\cite{chen2019three}}
\newcommand{\CPFP}{CPFP~\cite{zhao2019Contrast}}
\newcommand{\DMRA}{DMRA~\cite{piao2019depth}}
\newcommand{\DTNet}{D3Net~\cite{fan2019D3Net}}

\newcommand{\NJU}{\textit{NJU2K}~\cite{ju2014depth}}
\newcommand{\NLPR}{\textit{NLPR}~\cite{peng2014rgbd}}
\newcommand{\STERE}{\textit{STERE}~\cite{niu2012leveraging}}
\newcommand{\DES}{\textit{DES}~\cite{cheng2014depth}}
\newcommand{\SSD}{\textit{SSD}~\cite{zhu2017three}}
\newcommand{\SIP}{\textit{SIP}~\cite{fan2019D3Net}}

\begin{document}

\title{Bilateral Attention Network for RGB-D \\ Salient Object Detection}
%
%
%

\author{Zhao Zhang,
        Zheng Lin,
        Jun Xu,
        Wenda Jin,
        Shao-Ping Lu,
        and 
        Deng-Ping Fan

%
\thanks{Zhao Zhang (e-mail: zzhang@mail.nankai.edu.cn), Zheng Lin, Jun Xu, Shao-Ping Lu, and Deng-Ping Fan are with the TKLNDST, College of Computer Science, Nankai University. 
%
%
Wenda Jin is with Tianjin University.
Shao-Ping Lu is the corresponding author (e-mail: slu@nankai.edu.cn). 
}}

\maketitle

\begin{abstract}
Most existing RGB-D salient object detection (SOD) methods focus on the foreground region when utilizing the depth images.
However, the background also provides important information in traditional SOD methods for promising performance.
To better explore salient information in both foreground and background regions, this paper proposes a Bilateral Attention Network (BiANet) for the RGB-D SOD task.
Specifically, we introduce a Bilateral Attention Module (BAM) with a complementary attention mechanism: foreground-first (FF) attention and background-first (BF) attention.
The FF attention focuses on the foreground region with a gradual refinement style,
while the BF one recovers potentially useful salient information in the background region.
Benefitted from the proposed BAM module, our BiANet can capture more meaningful foreground and background cues,
and shift more attention to refining the uncertain details between foreground and background regions.
%
Additionally, we extend our BAM by leveraging the multi-scale techniques for better SOD performance.
Extensive experiments on six benchmark datasets demonstrate that our BiANet outperforms other state-of-the-art RGB-D SOD methods in terms of objective metrics and subjective visual comparison.
Our BiANet can run up to 80fps
on $224\times 224$ RGB-D images, with an NVIDIA GeForce RTX 2080Ti GPU.
Comprehensive ablation studies also validate our contributions.
%
\end{abstract}


\IEEEpeerreviewmaketitle

\section{Introduction}
\label{sec:intro}
\IEEEPARstart{S}{alient} object detection (SOD) aims to segment the most attractive objects in an image.
As an fundamental computer vision task, SOD has been widely applied into many vision applications, such as visual tracking~\cite{li2019gradnet,mahadevan2009saliency}, image segmentation~\cite{hou2018nips,jung2011unified,tsai2018image}, 
and video analysis~\cite{zhang2017study,RANet2019}, \textsl{etc}.
Most of existing SOD methods~\cite{jiang2019super,liu2019deep,zhang2020multistage} mainly deal with RGB images.
However, they usually produce inaccurate SOD results on the scenarios of similar texture, complex background, or homogeneous objects~\cite{wang2019focal,zhang2019salient}.
%
With the popularity of depth sensors in smartphones, the depth information, \eg, 3D layout and spatial cues, is crucial for reducing the ambiguity in the RGB images, and serves as important supplements to improve the SOD performance~\cite{liang2012depth}.
%

\begin{figure}[t]
	\centering
	\begin{overpic}[width=1\columnwidth]{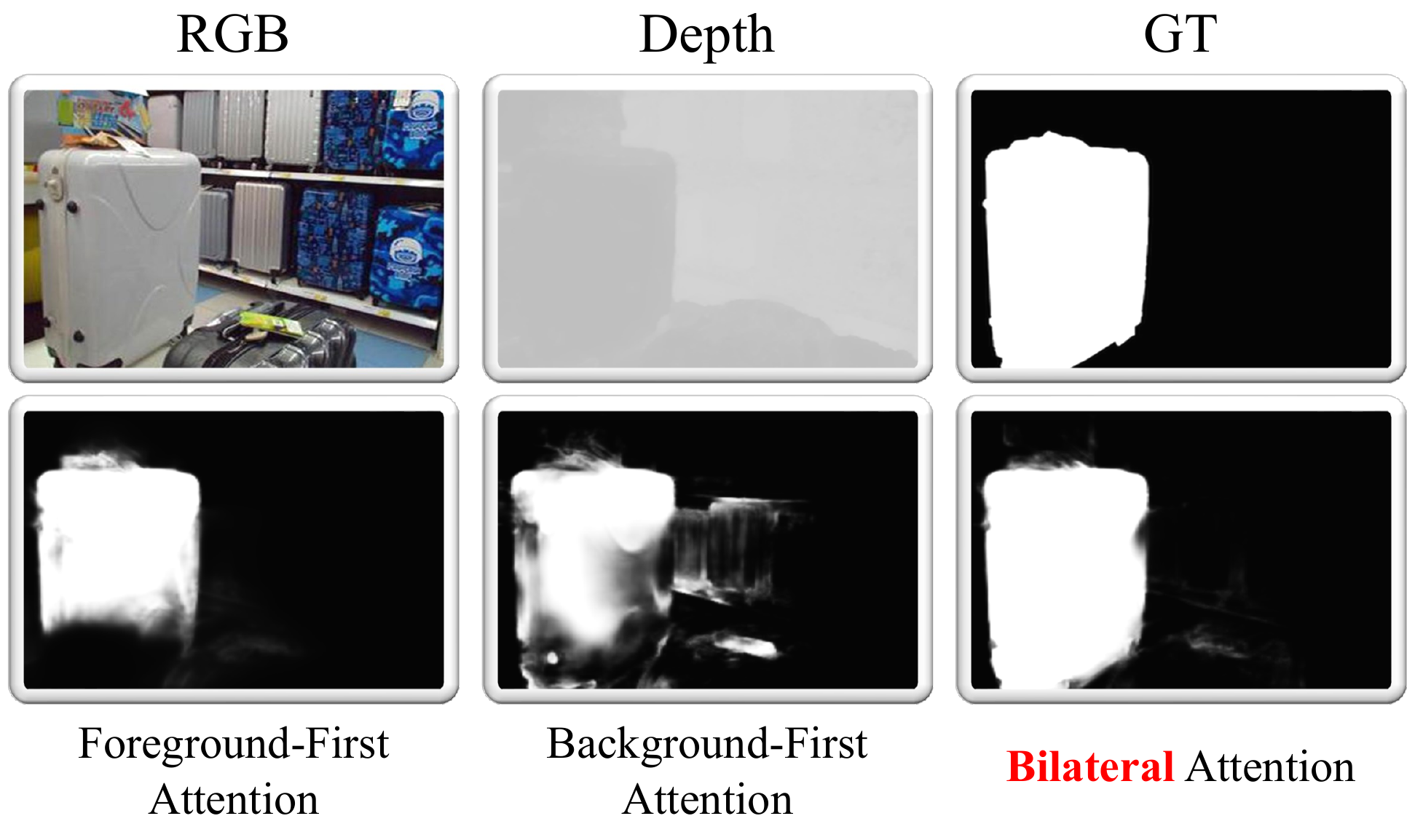} 
	\end{overpic}
	\vspace{-8mm}
	\caption{
	\textbf{Comparison of RGB-D SOD results by Foreground-First, Background-First, and our Bilateral attention mechanisms}.
	Depth information provides rich foreground and background relationships.
	Paying more attention to foreground helps to predict high-confidence foreground objects, but may produce incomplete results.
	Focusing more on background finds more complete objects, but may introduce unexpected noise.
	Our BiANet jointly explores foreground and background cues, and achieves complete foreground prediction with little background noise.
	}
	\vspace{-8pt}
	\label{fig:Motivation}
\end{figure}

\begin{figure*}[t]
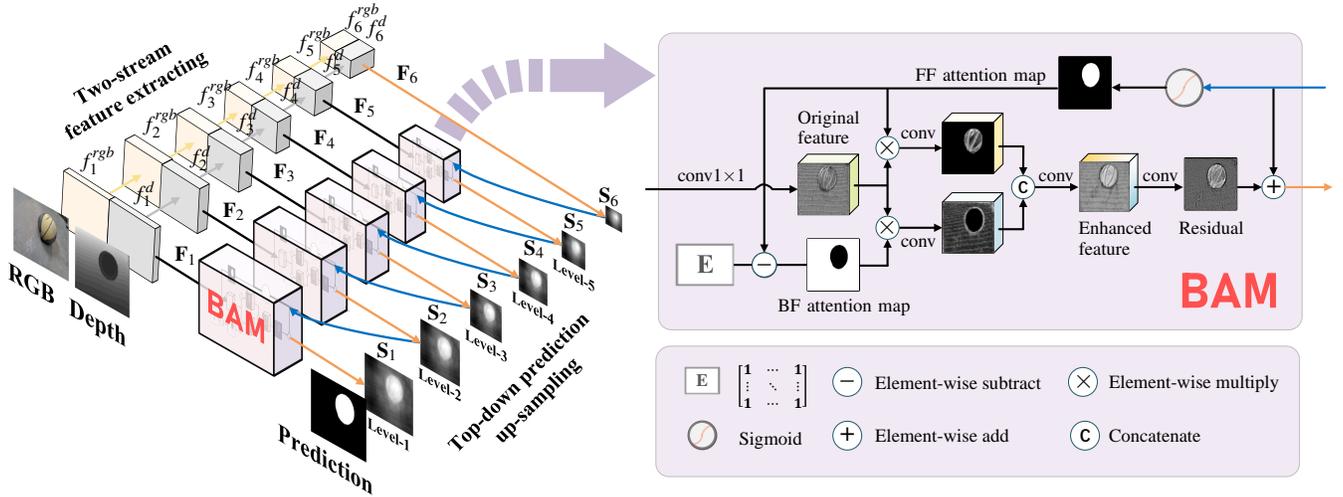

	\centering
	\begin{overpic}[width=.98\linewidth]{Pics/BiANet_TIP.pdf}
		\put(5.8, 25){\footnotesize{$f^{rgb}_1$}}
		\put(10.5, 27.8){\footnotesize{$f^{rgb}_2$}}
		\put(14.6, 30){\footnotesize{$f^{rgb}_3$}}
		\put(18.2, 31.9){\footnotesize{$f^{rgb}_4$}}
		\put(22, 33.9){\footnotesize{$f^{rgb}_5$}}
		\put(25.6, 35.7){\footnotesize{$f^{rgb}_6$}}
		
		\put(9.8, 22.6){\footnotesize{$f^{d}_1$}}
		\put(14, 25.3){\footnotesize{$f^{d}_2$}}
		\put(17.5, 28){\footnotesize{$f^{d}_3$}}
		\put(20.8, 30.3){\footnotesize{$f^{d}_4$}}
		\put(24, 32.5){\footnotesize{$f^{d}_5$}}
		\put(27.2, 35){\footnotesize{$f^{d}_6$}}
		
		\put(12.9, 18){\footnotesize{$\mathbf{F}_1$}}
		\put(16.5, 21.3){\footnotesize{$\mathbf{F}_2$}}
		\put(20.3, 24.2){\footnotesize{$\mathbf{F}_3$}}
		\put(23.3, 26.7){\footnotesize{$\mathbf{F}_4$}}
		\put(26.3, 29.1){\footnotesize{$\mathbf{F}_5$}}
		\put(29.5, 31.6){\footnotesize{$\mathbf{F}_6$}}
		
		\put(28, 10.8){\footnotesize{$\mathbf{S}_1$}}
		\put(31.8, 13.6){\footnotesize{$\mathbf{S}_2$}}
		\put(35.5, 16){\footnotesize{$\mathbf{S}_3$}}
		\put(39, 18.5){\footnotesize{$\mathbf{S}_4$}}
		\put(42, 20.5){\footnotesize{$\mathbf{S}_5$}}
		\put(44.6, 22.4){\footnotesize{$\mathbf{S}_6$}}
		
		\put(65.4, 20){$\times$}
		\put(65.4, 26){$\times$}
		\put(56.5, 17.25){--}
		\put(94.55, 23.05){+}
		
		\put(62.7, 8.6){--}
		\put(62.65, 4.55){+}
		\put(80.1, 8.5){$\times$}
		
		\put(50.5, 24){\scriptsize{conv1$\times$1}}
		\put(67.2, 27.2){\scriptsize{conv}}
		\put(67.2, 19.1){\scriptsize{conv}}
		\put(77.5, 24){\scriptsize{conv}}
		\put(85.3, 24){\scriptsize{conv}}
		
		\put(80.5, 20){\scriptsize{Enhanced}}
		\put(80.5, 18.5){\scriptsize{feature}}
		\put(88, 20){\scriptsize{Residual}}
		\put(68.3, 31.5){\scriptsize{FF attention map}}
		\put(58, 14.2){\scriptsize{BF attention map}}
		\put(59.5, 28.3){\scriptsize{Original}}
		\put(59.5, 26.8){\scriptsize{feature}}
	\end{overpic}
	\caption{\textbf{The overall architecture of our BiANet}.
		BAM denotes the proposed Bilateral Attention Module, and it also can be selectively replaced by its multi-scale extension (MBAM).
		BiANet contains three main steps: two-stream feature extracting,
		top-down prediction up-sampling, and bilateral attention residual compensation (by BAM).
		Specifically,
		it first extracts the multi-level features $\{f^{rgb}_i, f^d_i\}_{i=1}^6$ from the RGB and depth streams, and concatenates them to $\{\mathbf{F}_i\}_{i=1}^6$.
		We take the top feature $\mathbf{F}_6$ to predicate a coarse salient map $\mathbf{S}_6$.
		To obtain the accurate and high-resolution result,
		we up-sample the initial salient map and compensate the details by BAMs in a top-down manner.
		BAMs receive the higher-level prediction $\mathbf{S}_{i+1}$ and current level feature $\mathbf{F}_{i}$ as inputs. 
		In a BAM, the foreground-first attention map $\mathbf{A}^F_i$ and the background-first attention map $\mathbf{A}^B_i$ can be calculated according to $\mathbf{S}_{i+1}$.
		We apply the duel complementary attention maps to explore the foreground and background cues bilaterally, 
		and jointly infer the residual for refining the up-sampled saliency map.
	}
	\label{fig:Flow}
\end{figure*}

Recently, RGB-D SOD has received increasing research attention~\cite{chen2019three,piao2019depth}.
Early RGB-D SOD works~\cite{peng2014rgbd,ren2015exploiting,song2017depth} introduced the depth contrast as an important prior for the SOD task.
The recent work of CPFP~\cite{zhao2019Contrast} utilized the depth contrast prior to design an effectiveness loss.
These methods essentially explore depth information to shift more priority on the foreground region~\cite{chen2018progressively,chen2020improved}.
However, as demonstrated in~\cite{liang2018stereoscopic,xia2017what,xiao2018rgb}, understanding what background is can also promote the SOD performance.
Several traditional methods~\cite{li2015robust,yang2013saliency} predict salient objects jointly from the complementary foreground and background information, which is largely ignored by current RGB-D SOD networks.

In this paper, we propose a Bilateral Attention Network (BiANet) to collaboratively learn complementary foreground and background features from both RGB and depth streams for better RGB-D SOD performance.   
As shown in \figref{fig:Flow}, our BiANet employs a two-stream architecture, and the side outputs from the RGB and depth streams are concatenated in multiple stages.
Firstly, we use the high-level semantic features $\mathbf{F}_6$ to locate the foreground and background regions $\mathbf{S}_6$.
However, the initial saliency map $\mathbf{S}_6$ is coarse and in low-resolution.
To enhance the coarse saliency map, we design a Bilateral Attention Module (BAM), which is composed of the complementary foreground-first (FF) attention and background-first (BF) attention mechanisms.
The FF shifts attention on the foreground region to gradually refine its saliency prediction, while the BF focuses on the background region to recover the potential salient regions around the boundaries.
By bilaterally exploring the foreground and background cues,
the model helps predict more accurately
as shown in \figref{fig:Motivation}.
%
Secondly, we propose a multi-scale extension of BAM (MBAM) to effectively learn multi-scale contextual information, 
and capture both local and global saliency information to further improve the SOD performance.
Extensive experiments on six benchmark datasets demonstrate that our BiANet achieves better performance than previous state-of-the-arts on RGB-D SOD, and is very fast owing to our simple architecture.

In summary, our main contributions are three-fold:
\begin{itemize}
	\item \textbf{We propose a simple yet effective Bilateral Attention Module (BAM)} to explore the foreground and background cues collaboratively with the rich foreground and background information from the depth images.\
	\item \textbf{Our BiANet achieves better performance on six popular RGB-D SOD datasets} under nine standard metrics, and 
	presents better visual effects (\eg, contains more details and sharp edges) than the state-of-the-art methods.\
	\item \textbf{Our BiANet runs at 34fps$\sim$80fps} on an NVIDIA GeForce RTX2080Ti GPU under different settings, and is a feasible solution for real-world applications.
\end{itemize}

The remainder of this paper is organized as follows.
In \S\ref{sec:related}, we briefly survey the related work.
In \S\ref{sec:method}, we present the proposed Bilateral Attention Network (BiANet) for RGB-D Salient Object Detection.
Extensive experiments are conducted in \S\ref{sec:exp} to evaluate its performance when compared with state-of-the-art RGB-D SOD methods on six benchmark datasets.
The conclusion is given in \S\ref{conclusion}.

\begin{figure*}[t]
	\centering
	\begin{overpic}[width=.96\textwidth]{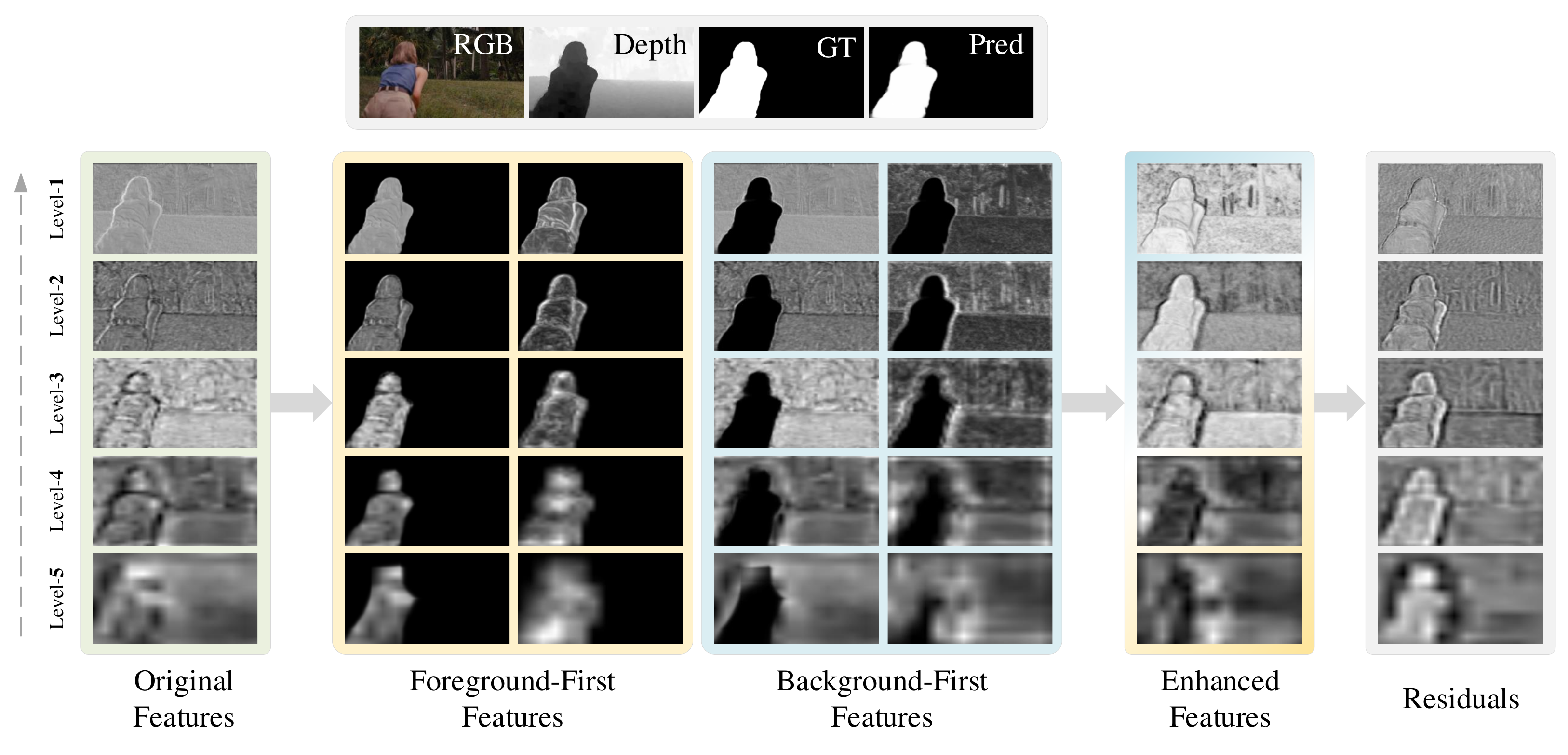}
	\end{overpic}
	\caption{\textbf{Visualizing the working mechanism of bilateral attention}.
		The original features are the averaged side-output features in each levels.
		We show the original features directly multiplied by foreground- and background-first attention maps in left columns of yellow and blue boxes.
		The right columns of the two boxes are the further convoluted features in two branches.
		As can be seen, 
		the foreground-first features focus on foreground region to explore the saliency cues;
		while the background-first features shift more attention to the background regions to mine the potentially significant objects.
		No matter in the features of foreground- or background-first features, 
		more priority is shifted to the uncertain areas caused by the up sampling.
		When fusing the two branches and jointly inferring,
		we can see the bilaterally enhanced features have a more accurate understanding
		where the foreground or background is.
		Due to obtaining more attention, 
		the uncertain areas are reassigned to the right attribution by the residual with strong contrast.
		'Pred' is the prediction of the model.
	}
	\label{fig:VisBAM}
\end{figure*}

\section{Related Work}
\label{sec:related}

\subsection{RGB-D Salient Object Detection}
RGB-D salient object detection (SOD) aims to segment the most attractive object(s) in a pair of cross-modal RGB and depth images.\
Early methods mainly focus on extracting low-level saliency cues from RGB and depth images, exploring object distance~\cite{liang2012depth}, difference of Gaussian~\cite{ju2014depth}, graph knowledge~\cite{cong2016saliency}, multi-level discriminative saliency fusion~\cite{song2017depth}, multi-contextual contrast~\cite{cong2019going,peng2014rgbd}, and background enclosure~\cite{feng2016local}, \etc.
However, these methods often produce inaccurate saliency predictions, due to the lack of high-level feature representation.

Recently, deep neural networks (DNNs)~\cite{he2016deep} have been employed to investigate high-level representations of cross-modal fusion of RGB and depth images, with much better SOD performance.\ 
Most of these DNNs~\cite{chen2019multi,han2017cnns,wang2019adaptive} first extract the RGB and depth features separately and then fuse them in the shallow, middle, or deep layers of the network.\  
The methods of~\cite{chen2018progressively,chen2019three,li2020icnet,piao2019depth} further improved the SOD performance by fusing cross-modal features in multi-level stages instead of as a one-off integration.\
Zhao \etal\cite{zhao2019Contrast} also took the enhanced depth image as attention maps to boost RGB features in multiple stages with better SOD performance.

\subsection{Foreground and Background Cues}
\vspace{-3pt}
There are great differences in the distribution of foreground and background, so it is necessary to explore their respective cues.\
In traditional methods, some works focus on reasoning salient areas in foreground and background jointly.\ 
Yang \etal\cite{yang2013saliency} proposed a two-stage method for SOD.\  
It first takes the four boundaries of the inputs as background seeds to infer foreground queries via a graph-based manifold ranking.\ 
Then, it ranks the graph depending on the foreground seeds in the same manner for final detection.\ 
This method is enlightening, but it has obvious limitations: 1) It is inappropriate to use the four boundaries directly as background, because the foreground is likely to be connected to the boundaries.\ 
2) Aggregation at the super-pixel level also results in rough outputs.\ 
For the limitation 1), 
Liang~\etal\cite{liang2018stereoscopic} introduce the depth map to distinguish foreground and background regions instead of only assuming the boundaries as background. The depth map shows clear disparity in most senses; thus, it can support more precise locating.
For the limitaion 2), Li \etal\cite{li2015robust} further used the regularized random walks ranking to formulate pixel-wised saliency maps, which improves the scaling effect caused by super-pixel aggregation.\ 
Nevertheless, only depending on these low-level priors, traditional methods cannot accurately locate the initial region of foreground and background.  

Recently, Chen \etal\cite{chen2018reverse} proposed to gradually explore saliency regions from the background using reverse attention, but they ignored the contribution of foreground cues to the final detection.\ 
As far as we know, how to jointly refine the salient objects from the foreground and background regions is still an open problem in deep RGB-D SOD methods.

\section{Proposed BiANet for RGB-D SOD}
\label{sec:method}
In this section, we first introduce the overall architecture of our BiANet, and then present the bilateral attention module (BAM) as well as its multi-scaled extension (MBAM).

\subsection{Architecture Overview}
As shown in \figref{fig:Flow}, our Bilateral Attention Network (BiANet) contains three main steps: feature extracting, prediction up-sampling, and bilateral attention residual compensation.\
We extract the multi-level features from the RGB and depth streams.\
With increasing network depth, the high-level features (\eg, $\mathbf{F}_4$) will be more potent for capturing global context,
while it loses the object details.\
When we up-sample the high-level predictions, the saliency maps (\eg, $\mathbf{S}_5$) will be blurred, \eg the edges will become uncertain.\
Thus, we use the proposed Bilateral Attention Module (BAM) to distinguish foreground and background regions.\

\subsubsection{Feature extracting}
We encode RGB and depth information with two streams.\
Specifically, both the RGB and depth streams employ five convolutional blocks from VGG-16~\cite{simonyan2015vgg} as the standard backbone
and attach an additional convolution group with three convolutional layers to predict the saliency maps, respectively.\
Unlike previous works~\cite{han2017cnns,zhu2019pdnet,chen2019multi},
we explore the cross-modal fusion of RGB and depth features at multiple stages, rather than fusing them once in low or high stage.\
The $i$-th side output $f_i^{rgb}$ from the RGB stream and $f_i^{d}$ from the depth stream are concatenated as a feature tensor $\mathbf{F}_i$.\
Note that, $\mathbf{F}_6$ is concatenated by $M(f_5^{rgb})$ and $M(f_5^{d})$, where $M(\cdot)$ denotes the max-pooling operation.\
The coarse saliency map $\mathbf{S}_6$ is derived from $\mathbf{F}_6$, 
and $\{\mathbf{F}_1,\mathbf{F}_2,\cdots,\mathbf{F}_5\}$ are prepared for the BAMs in our BiANet to further refine the up-sampled saliency maps, by distinguishing the uncertain regions as foreground or background in a top-down manner.

\subsubsection{Prediction up-sampling}
The initial saliency map predicted from the high-level features is coarse in low-resolution, but useful to predict the initial position of the foreground and background, since it contains rich semantic information.
To refine the basic saliency map $\mathbf{S}_6$, a lower-level feature $\mathbf{F}_5$ with more details is used to predict the residual component between the higher-level prediction and the ground-true ($\mathbf{GT}$) with the help of BAM.
We add the predicted residual component $\mathbf{R}_5$ to the up-sampled higher-level prediction $S_6$, and obtain a refined prediction $\mathbf{S}_5$, \etc., that is,
\begin{equation}
\mathbf{S}_i = \mathbf{R}_i + U(\mathbf{S}_{i+1}), i\in\{1,\dots,5\},
\label{equ:up}
\end{equation}
where $U(\cdot)$ means up-sampling.
Finally, our BiANet obtains a saliency map by $\mathbf{S}=\sigma(\mathbf{S}_1)$, where $\sigma(\cdot)$ is a sigmoid function.

\begin{figure}[t]
	\centering
	\begin{overpic}[width=.98\columnwidth]{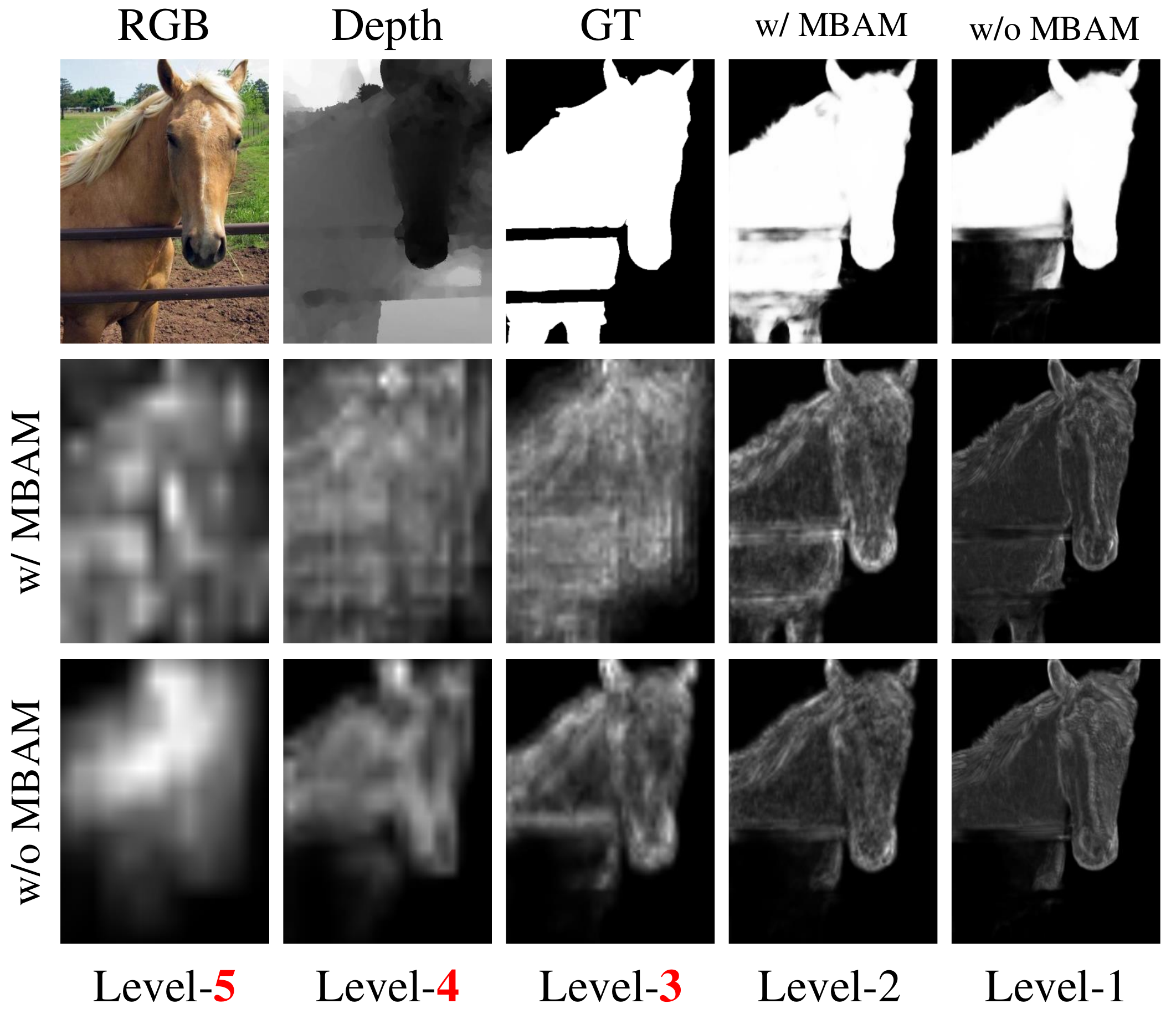} 
	\end{overpic}
	\caption{
		\textbf{Comparison of the high-level features capured by MBAM and BAM.}
		The second row is the averaged foreground-first features from the model where the MBAMs are applied in the top three levels (marked with red numbers).
		The thrid row is the averaged foreground-first features obtained from the model in which all levels are armed with BAMs.
		We can see that, compared with applying the BAMs, MBAMs in higher levels capture more complete information, 
		which is conducive to the object locating as shown in the first row.
	}
	\vspace{-8pt}
	\label{fig:VisMBAM}
\end{figure}

\subsubsection{Bilateral attention residual compensation}
To get better residuals and distinguish up-sampled foreground and background regions, we design a bilateral attention module (BAM) to enable our BiANet to discriminate the foreground and background.
In our BAM, the higher-level prediction serves as a foreground-first attention (FF) map, 
and the reversed prediction serves as background-first (BF) attention map to combine the bilateral attention on foreground and background.
In \figref{fig:VisBAM}, one can see that the residual generated by BAM possesses high contrast at the object boundaries.
More details are described in Sections \ref{sec:BAM} and \ref{sec:MBAM}.

\subsubsection{Loss function}
Deep supervision is widely used in the SOD task~\cite{feng2019attentive,hou2019deeply}.
It clarifies the optimization goals for each step of the network, and accelerates the convergence of training.
For quick convergence,
we also apply deep supervision in the depth stream output $\mathbf{S}_{d}$, 
RGB stream output $\mathbf{S}_{rgb}$, and each top-down side output $\{\mathbf{S}_1,\mathbf{S}_2,\cdots,\mathbf{S}_6\}$.
The total loss function of our BiANet is
\begin{equation}
\begin{split}
\mathcal{L} = & \sum\nolimits_{i=1}^6w_i\mathcal{L}_{ce}\left(\sigma\left(\mathbf{S}_i\right), \mathbf{GT}\right) + w_d\mathcal{L}_{ce}\left(\sigma\left(\mathbf{S}_{d}\right), \mathbf{GT}\right) \\
&+w_{rgb}\mathcal{L}_{ce}\left(\sigma\left(\mathbf{S}_{rgb}\right), \mathbf{GT}\right),
\end{split}
\end{equation}
in which $w_i, w_d,$ and $w_{rgb}$ are the weight coefficients
and simply set to $1$ in our experiments. $\mathcal{L}_{ce}(\cdot)$ is the binary cross entropy loss, which is formulated as
\begin{equation}
\mathcal{L}_{ce}(\mathbf{X}, \mathbf{Y})\!=\! -\frac{1}{N} \sum\limits_{i=1}^N\Big(y_{i}log(x_{i}) + (1 - y_{i})log(1-x_{i})\Big).
\end{equation}
In the above equation, $x_i\in \mathbf{X}$ and $y_i\in \mathbf{Y}$, and $N$ denotes the total pixel number.

\subsection{Bilateral Attention Module (BAM)}\label{sec:BAM}
\vspace{-3pt}
Given the initial foreground and background,
how to refine the prediction using higher-resolution cross-modal features is the focus of this paper.
Considering that the distribution of foreground and background are quite different,
we design a bilateral attention module using a pair of reversed attention components to
learn features from the foreground and background respectively, and then jointly refine the prediction.
As can be seen in \figref{fig:Flow}, 
to focus more on the foreground,
we use the up-sampled prediction from the higher-level as foreground-first attention (FF) maps $\{\mathbf{A}^F\}_{i=1}^5$ after they are activated by sigmoid,
and the background-first attention (BF) maps $\{\mathbf{A}^B\}_{i=1}^5$ are generated by subtracting FF maps from matrix $\mathbf{E}$, in which all the elements are $1$.
\begin{equation}
\left\{
\begin{array}{l}
\mathbf{A}^F_i=\sigma \Big(U(\mathbf{S}_{i+1})\Big), \\
\mathbf{A}^B_i=\mathbf{E} - \sigma \Big(U(\mathbf{S}_{i+1})\Big), \\
\end{array}
\right.
i\in\{1,2,3,4,5\}.
\end{equation}
Then, as shown in \figref{fig:Flow}, we apply FF and BF to weight the side-output features in two branches, respectively, and further predict the residual component jointly.
\begin{equation}
\mathbf{R}_i = \mathcal{P}_R\left(\left[\mathcal{P}\left( \hat{\mathbf{F}}_i\odot \mathbf{A}^F_i\right),\mathcal{P}\left(\hat{\mathbf{F}}_i\odot \mathbf{A}^B_i\right)\right]\right),
\end{equation}
where $\hat{\mathbf{F}}_i$ is the channel-reduced feature of ${\mathbf{F}}_i$ using 32 $1\times1$ convolutions to reduce the computational cost.
$\mathcal{P}$ represents the feature extraction operation consisting of 32 convolution kernels with a size of $3\times3$ and a ReLU layer.
The two branches do not share parameters.
$[\cdot,\cdot]$ means concatenation.
$\mathcal{P}_R$ is the prediction layer to output a single channel residual map via a $3\times3$ kernel after the same feature extraction operation with $\mathcal{P}$.
Once the $\mathbf{R}_i$ is obtained, the refined prediction $\mathbf{S}_i$ is obtained via \equref{equ:up}.

To better understand the working mechanism of BAM, in \figref{fig:VisBAM}, we visualize the channel-wise averaged features from BAMs in different levels.
In BAM, the original features will be first fed into two branches by multiply the FF and BF attention maps, respectively.
The result of the direct multiplication is illustrated in the left half of the yellow (FF features) and blue (BF features) boxes.
We can see that FF branch shifts attention to the foreground area predicted from its higher level to explore foreground saliency cues.
After a convolution layer, more priority is given to the uncertain area.
Complementarily, BF branch focuses on the background area to explore the background cues, looking for possible salient objects within it.
%
%
In our BiANet,
the top-down prediction up-sampling is a process in which the resolution of salient objects is gradually increased. 
It will result in uncertain coarse edges.
We can see that both of FF and BF features focus on the uncertain area (such as object boundaries).
The low-level and high-resolution FF branch will eliminate the overflow of the uncertain area, 
while the BF branch will eliminate the uncertain area which does not belong to the background.
That is an important reason why BiANet performs better on detail
and is prone to predicting sharp edges.
After the joint inferring, we can see the bilaterally enhanced features contain more discriminative spatial information of foreground and background.
The generated residual components are with sharp contrast on the edges,
and then suppress the background area and strengthen the foreground regions.
\begin{figure*}[t!]
	\centering
	\footnotesize
	\begin{minipage}[b]{0.3\linewidth}
		\begin{overpic}[width=.98\linewidth]{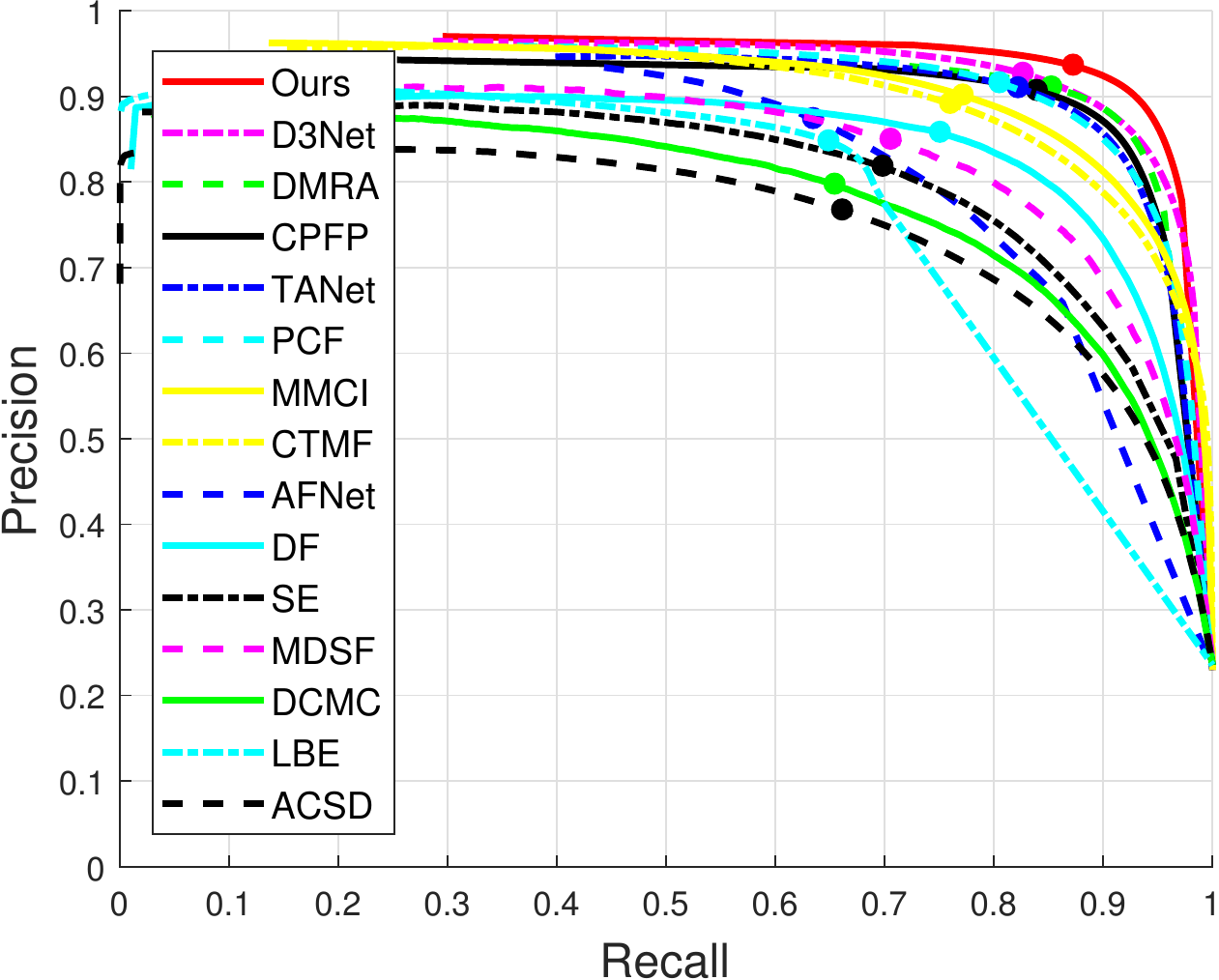}
			\put(35,35){\Large{\NJU}}
		\end{overpic}
		\vspace{1pt}
	\end{minipage}
	\begin{minipage}[b]{0.3\linewidth}
		\begin{overpic}[width=.98\linewidth]{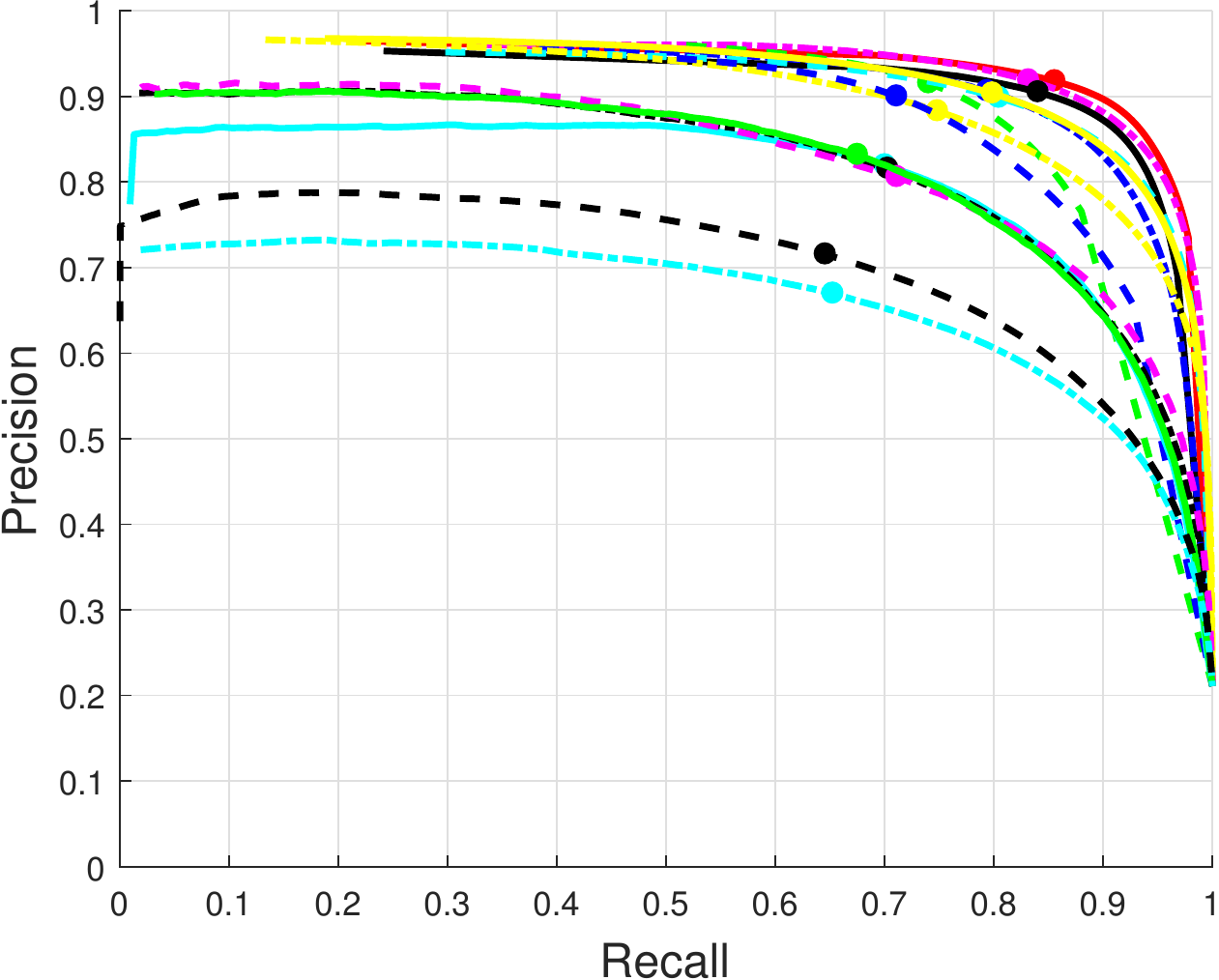}
			\put(35,35){\Large{\STERE}}
		\end{overpic}
		\vspace{1pt}
	\end{minipage}
	\begin{minipage}[b]{0.3\linewidth}
		\begin{overpic}[width=.98\linewidth]{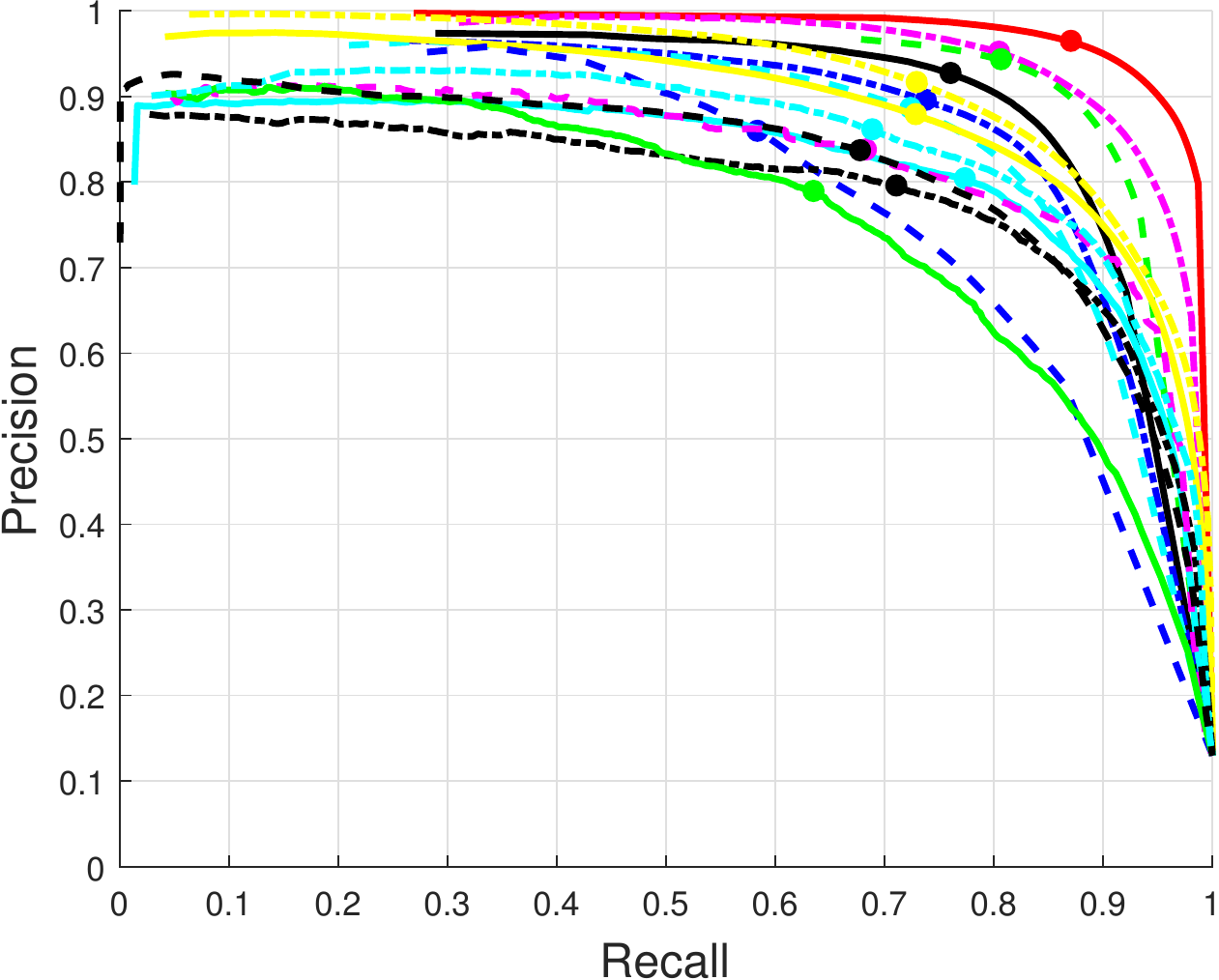}
			\put(35,35){\Large{\DES}}
		\end{overpic}
		\vspace{1pt}
	\end{minipage}

	\begin{minipage}[b]{0.3\linewidth}
		\begin{overpic}[width=.98\linewidth]{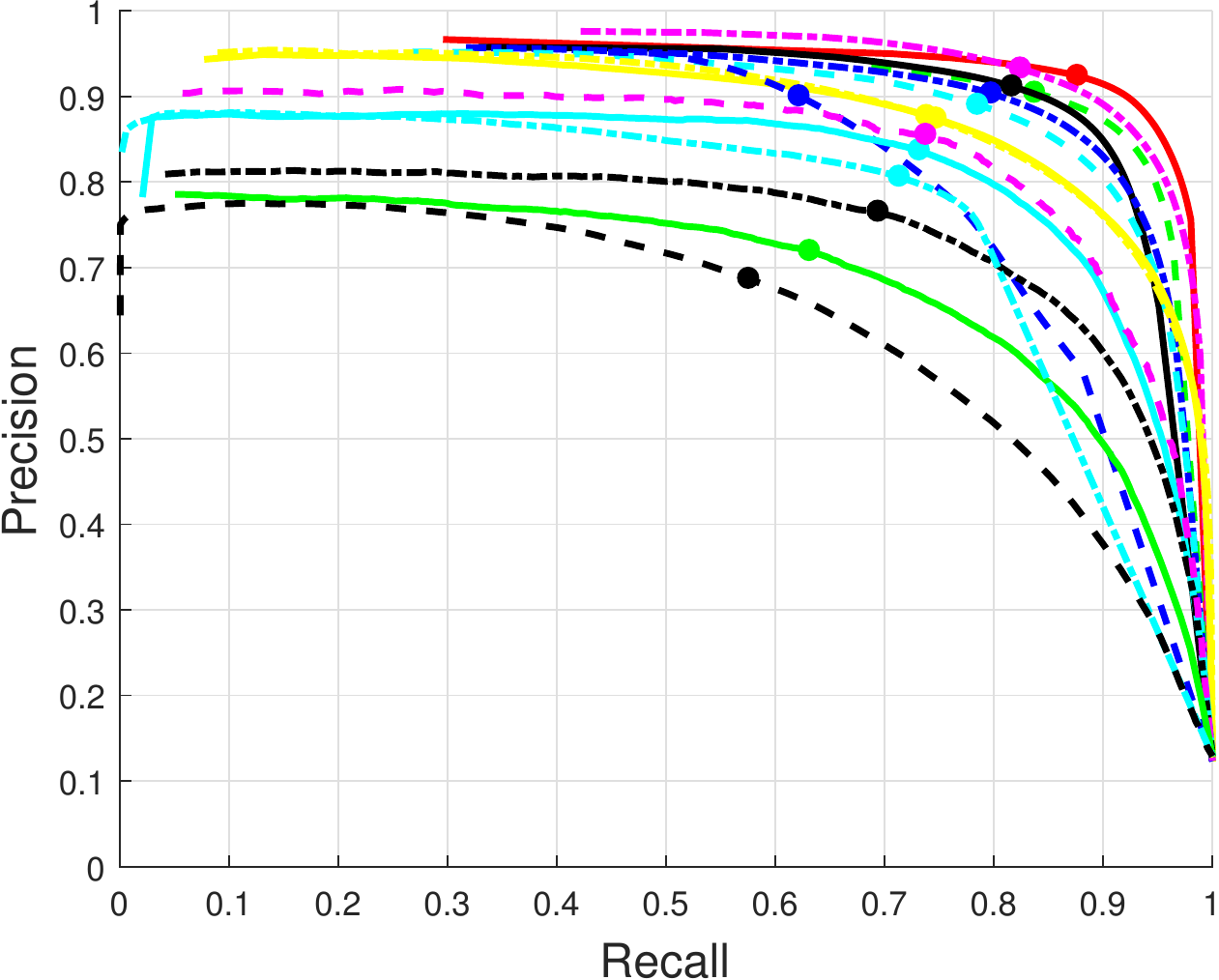}
			\put(35,35){\Large{\NLPR}}
		\end{overpic}
	\end{minipage}
	\begin{minipage}[b]{0.3\linewidth}
		\begin{overpic}[width=.98\linewidth]{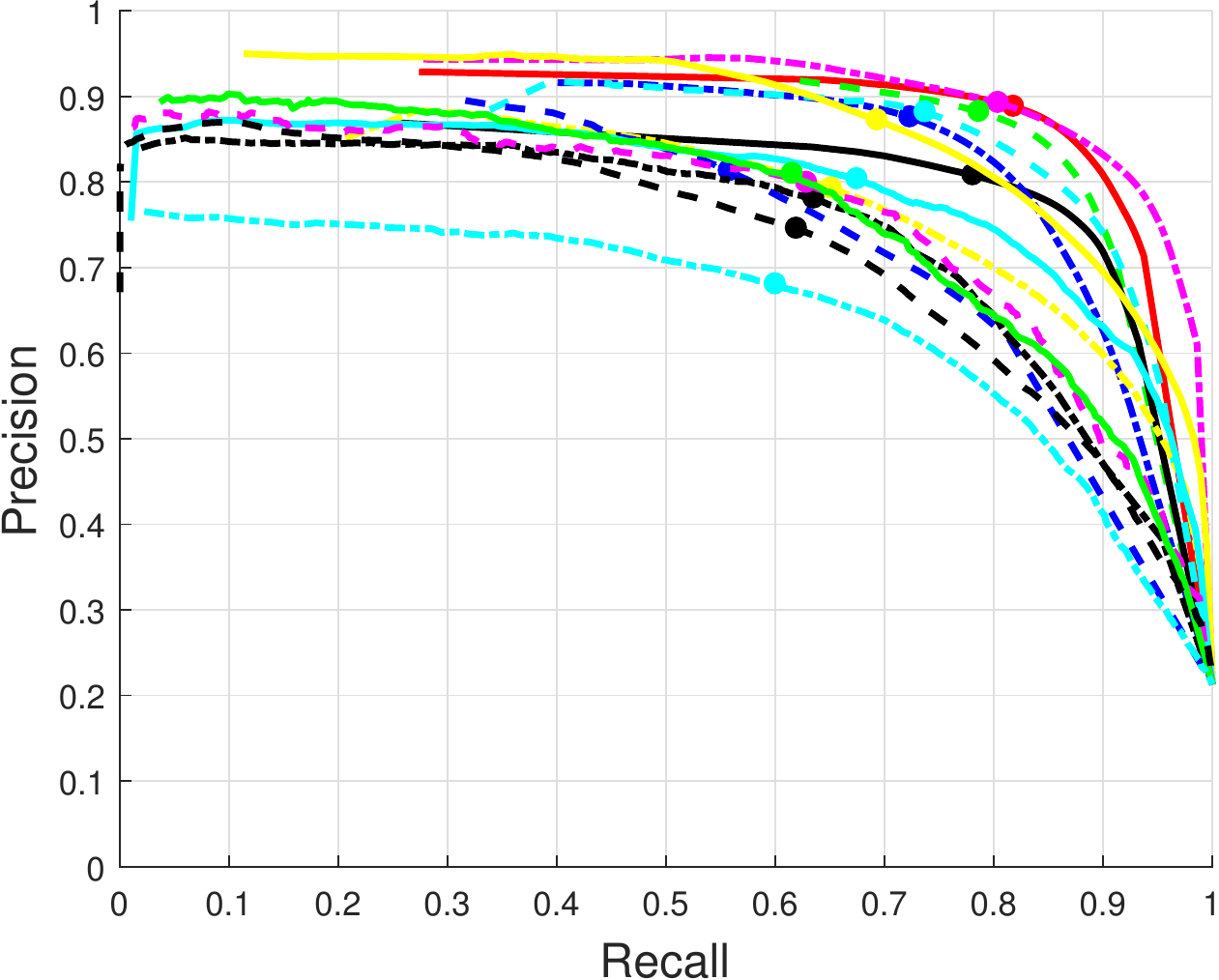}
			\put(35,35){\Large{\SSD}}
		\end{overpic}
	\end{minipage}
	\begin{minipage}[b]{0.3\linewidth}
		\begin{overpic}[width=.98\linewidth]{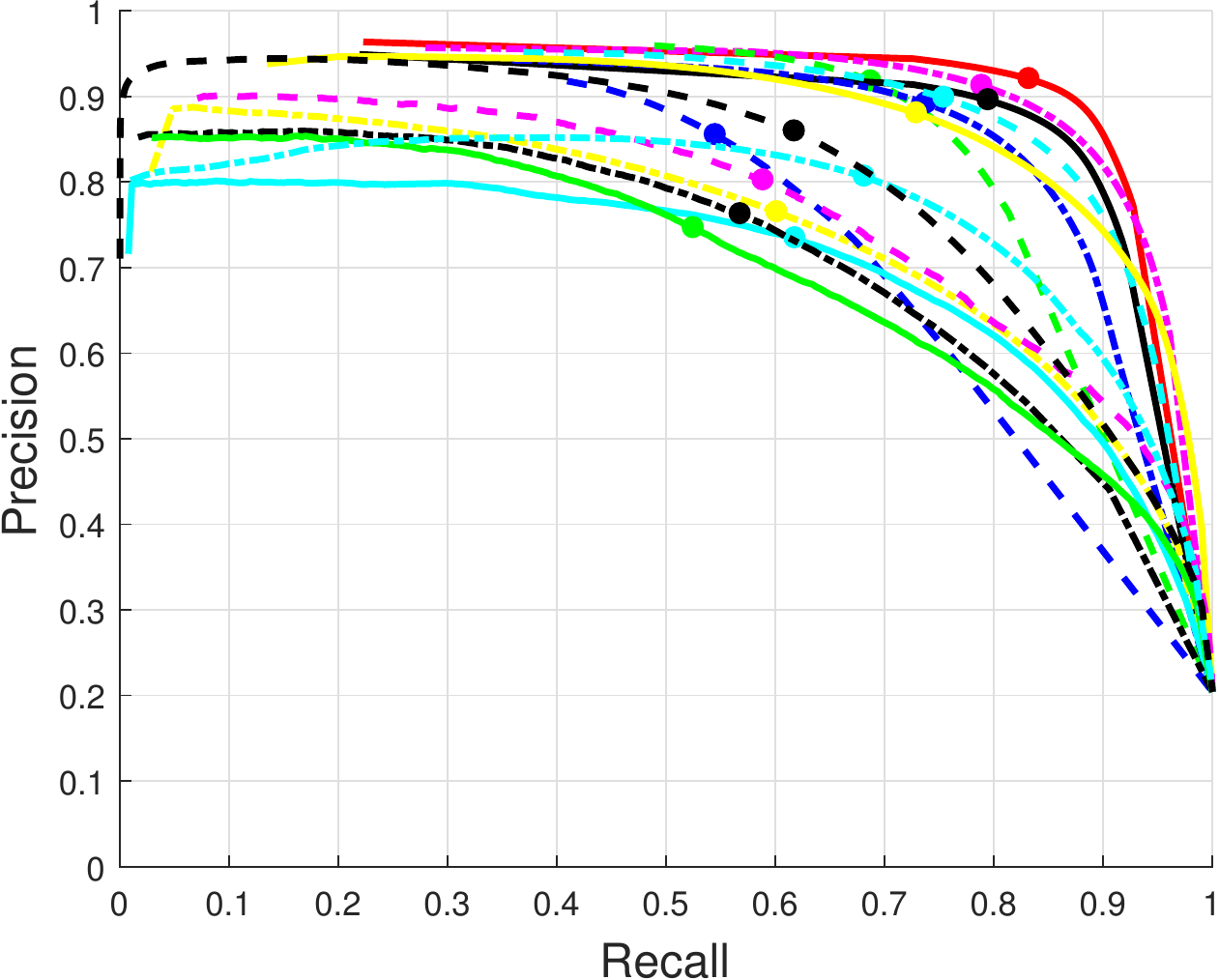}
			\put(35,35){\Large{\SIP}}
		\end{overpic}
	\end{minipage}
	%
	\caption{\textbf{PR curves of our BiANet and other 14 state-of-the-art methods across 6 datasets}. 
		The node on each curve denotes the precision and recall value used for calculating max F-measure.}
	\label{fig:PR}
\end{figure*}

		
		

\subsection{Multi-Scale Extension of BAM (MBAM)}\label{sec:MBAM}
Salient objects in a scene are various in location, size, and shape.
Thus, exploring the multi-scaled context in high-level layers benefits for understanding the scene~\cite{wang2019salinet,zhao2019pyramid}.
To this end, we extend our BAM with a multi-scale version, in which groups of dilated convolutions are used to extract pyramid representations from the undetermined foreground and background areas.
Specifically, the module can be described as
\begin{equation}
\mathbf{R}_i\ = \mathcal{P}_R\Big(\left[\sqcup_{i=1}^4\mathcal{D}_i\left( \mathbf{F}_i\odot\mathbf{A}^F_i\right),\sqcup_{i=1}^4\mathcal{D}_i\left(\mathbf{F}_i\odot \mathbf{A}^B_i\right)\right]\Big),
\end{equation}
where $\sqcup$ means a concatenate operation.
$\mathcal{D}_1$ consists of $1\times1$ kernels with 32 channels and a ReLU layer.
$\{\mathcal{D}_i\}_{i=2}^4$ is a group of dilated convolutions, with rates of 3, 5, and 7.
They all consist of $3\times3$ kernels with 32 channels and a ReLU layer.

We recommend applying the MBAM in high-level cross-modal features, such as $\{\mathbf{F}_3,\mathbf{F}_4,\mathbf{F}_5\}$,
which need different sizes of receptive fields to explore multi-scale context.
MBAM effectively improves the detection performance but introduces a certain computational cost.
Thus, the number of MBAM should be a trade-off in practical applications. 
In \secref{sec:abl_mbam}, we discuss in detail how the number of MBAM changes the detection effect and calculation cost.

In order to intuitively observe the gain effect brought by MBAM,
we visualize the averaged foreground-first feature maps from MBAMs and BAMs in \figref{fig:VisMBAM}.
In the second row,
the feature maps are obtained from the model with three MBAMs in its top three levels,
while in the last row, 
all the feature maps are collected from BAMs. 
We can see the target object (horse) account for a large proportion of the scene.
Without the ability to perceive multi-scale information effectively,
the BAM fails to capture the accurate global salient regions in high levels
and leads to incomplete prediction finally.
When introducing the multi-scale extension, 
we can see higher-level features achieve stronger spatial representation,
which supports to locate more complete salient object.

\subsection{Implementation Details}
\subsubsection{Settings}
We apply the MBAM in the high-level side outputs $\{\mathbf{F}_3,\mathbf{F}_4,\mathbf{F}_5\}$ during implementation,
and use bilinear interpolation in all interpolation operations.
The initial parameters of our backbone are loaded from a VGG-16 network pre-trained on ImageNet.
Our BiANet is based on PyTorch~\cite{pytorch2019paszke}.


\subsubsection{Training}
Following \DTNet, we use the training set containing 1485 and 700 image pairs from the \textit{NJU2K}~\cite{ju2014depth}
and \textit{NLPR}~\cite{peng2014rgbd} datasets, respectively.
We employ the Adam optimizer~\cite{kingma2015adam} with an initial learning rate of 0.0001, $\beta_1=0.9$, and $\beta_2=0.99$.
The batch size is set to 8,
and we train our BiANet for 25 epochs in total.
The training images are resized to $224 \times 224$, also during the test.
The output saliency maps are resized back to the original size for evaluation.
Accelerated by an NVIDIA GeForce RTX 2080Ti, 
our BiANet takes about 2 hours for training, 
and runs at 34$\sim$80fps (with different numbers of MBAMs) for the inputs with $224\times224$ resolution.

\begin{table*}[t!]
	\centering
	\caption{
		Quantitative comparisons of our BiANet with nine deep-learning-based methods and five traditional methods on six popular datasets in term of S-measure ($S_{\alpha}$), maximum F-measure (max $F_{\beta}$), mean F-measure (mean $F_{\beta}$), adaptive F-measure (adp $F_{\beta}$), maximum E-measure (max $E_{\xi}$), mean E-measure (mean $E_{\xi}$), adaptive E-measure (adp $E_{\xi}$), and mean absolute error (MAE, $\mathcal{M}$).
		$F_{\beta}$ and $E_{\xi}$ represent max $F_{\beta}$ and max $E_{\xi}$ by default.
		$\uparrow$ means that the larger the numerical value, the better the model,
		while $\downarrow$ means the opposite.
		For traditional methods, the statistics are based on overall datasets rather on the test set.
	}
	\renewcommand{\arraystretch}{0.9}
	\renewcommand{\tabcolsep}{1.15mm}
	\begin{tabular}{lr|ccccc|ccccccccc|c}
		\hline\toprule
		
		&  & ACSD & LBE & DCMC & MDSF   & SE   & DF   & AFNet& CTMF & MMCI & PCF   & TANet& CPFP & DMRA & D3Net & BiANet \\
		& Metric & \scriptsize{ICIP14} & \scriptsize{CVPR16} & \scriptsize{SPL16} & \scriptsize{TIP17}   & \scriptsize{ICME16}   & \scriptsize{TIP17}   & \scriptsize{arXiv19} & \scriptsize{TOC18} & \scriptsize{PR19} & \scriptsize{CVPR18}   & \scriptsize{TIP19} & \scriptsize{CVPR19} & \scriptsize{ICCV19} & \scriptsize{arXiv19} & \scriptsize{2020} \\
		&  & \cite{ju2014depth} & \cite{feng2016local} & \cite{cong2016saliency} & \cite{song2017depth} & \cite{guo2016salient} & \cite{qu2017rgbd} & \cite{wang2019adaptive} & \cite{han2017cnns} & \cite{chen2019multi} & \cite{chen2018progressively} & \cite{chen2019three}& \cite{zhao2019Contrast} & \cite{piao2019depth} & \cite{fan2019D3Net}& Ours\\
		\hline
		\multirow{8}{*}{\begin{sideways}\NJU\end{sideways}}
		& $S_{\alpha}\uparrow$    & 0.699 & 0.695 & 0.686 & 0.748 & 0.664 & 0.763 & 0.772 & 0.849 & 0.858 & 0.877 & 0.878 & 0.879 & 0.886 & \tbb{0.893} & \trb{0.915} \\
		& max $F_{\beta}\uparrow$     & 0.711 & 0.748 & 0.715 & 0.775 & 0.748 & 0.804 & 0.775 & 0.845 & 0.852 & 0.872 & 0.874 & 0.877 & 0.886 & \tbb{0.887} & \trb{0.920} \\
		& mean $F_{\beta}\uparrow$     & 0.512 & 0.606 & 0.556 & 0.628 & 0.583 & 0.650 & 0.764 & 0.779 & 0.793 & 0.840 & 0.841 & 0.850 & \tbb{0.873} & 0.859 & \trb{0.903} \\
		& adp $F_{\beta}\uparrow$     & 0.696 & 0.740 & 0.717 & 0.757 & 0.734 & 0.784 & 0.768 & 0.788 & 0.812 & 0.844 & 0.844 & 0.837 & \tbb{0.872} & 0.840 & \trb{0.892} \\
		& max $E_{\xi}\uparrow$       & 0.803 & 0.803 & 0.799 & 0.838 & 0.813 & 0.864 & 0.853 & 0.913 & 0.915 & 0.924 & 0.925 & 0.926 & 0.927 & \tbb{0.930} & \trb{0.948} \\
		& mean $E_{\xi}\uparrow$     & 0.593 & 0.655 & 0.619 & 0.677 & 0.624 & 0.696 & 0.826 & 0.846 & 0.851 & 0.895 & 0.895 & 0.910 & \tbb{0.920} & 0.910 & \trb{0.934} \\
		& adp $E_{\xi}\uparrow$     & 0.786 & 0.791 & 0.791 & 0.812 & 0.772 & 0.835 & 0.846 & 0.864 & 0.878 & 0.896 & 0.893 & 0.895 & \tbb{0.908} & 0.894 & \trb{0.926} \\
		& $\mathcal{M}\downarrow$ & 0.202 & 0.153 & 0.172 & 0.157 & 0.169 & 0.141 & 0.100 & 0.085 & 0.079 & 0.059 & 0.060 & 0.053 & \tbb{0.051} & \tbb{0.051} & \trb{0.039} \\
		\midrule
		\multirow{8}{*}{\begin{sideways}\STERE\end{sideways}}
		& $S_{\alpha}\uparrow$    & 0.692 & 0.660 & 0.731 & 0.728 & 0.708 & 0.757 & 0.825 & 0.848 & 0.873 & 0.875 & 0.871 & 0.879 & 0.835 & \tbb{0.889} & \trb{0.904} \\
		& max $F_{\beta}\uparrow$    & 0.669 & 0.633 & 0.740 & 0.719 & 0.755 & 0.757 & 0.823 & 0.831 & 0.863 & 0.860 & 0.861 & 0.874 & 0.847 & \tbb{0.878} & \trb{0.898} \\
		& mean $F_{\beta}\uparrow$     & 0.478 & 0.501 & 0.590 & 0.527 & 0.610 & 0.617 & 0.806 & 0.758 & 0.813 & 0.818 & 0.828 & \tbb{0.841} & 0.837 & \tbb{0.841} & \trb{0.879} \\
		& adp $F_{\beta}\uparrow$     & 0.661 & 0.595 & 0.742 & 0.744 & 0.748 & 0.742 & 0.807 & 0.771 & 0.829 & 0.826 & 0.835 & 0.830  & \tbb{0.844} & 0.829 & \trb{0.873} \\
		& max $E_{\xi}\uparrow$      & 0.806 & 0.787 & 0.819 & 0.809 & 0.846 & 0.847 & 0.887 & 0.912 & 0.927 & 0.925 & 0.923 & 0.925 & 0.911 & \tbb{0.929} & \trb{0.942} \\
		& mean $E_{\xi}\uparrow$     & 0.592 & 0.601 & 0.655 & 0.614 & 0.665 & 0.691 & 0.872 & 0.841 & 0.873 & 0.887 & 0.893 & \tbb{0.912} & 0.879 & 0.906 & \trb{0.926} \\
		& adp $E_{\xi}\uparrow$     & 0.793 & 0.749 & 0.831 & 0.830 & 0.825 & 0.838 & 0.886 & 0.864 & 0.901 & 0.897 & \tbb{0.906} & 0.903 & 0.900 & 0.902 & \trb{0.926} \\
		& $\mathcal{M}\downarrow$ & 0.200 & 0.250 & 0.148 & 0.176 & 0.143 & 0.141 & 0.075 & 0.086 & 0.068 & 0.064 & 0.060 & \tbb{0.051} & 0.066 & 0.054 & \trb{0.043} \\
		\midrule
		\multirow{8}{*}{\begin{sideways}\DES\end{sideways}}
		& $S_{\alpha}\uparrow$    & 0.728 & 0.703 & 0.707 & 0.741 & 0.741 & 0.752 & 0.770 & 0.863 & 0.848 & 0.842 & 0.858 & 0.872 & \tbb{0.900} & 0.898 & \trb{0.931} \\
		& max $F_{\beta}\uparrow$     & 0.756 & 0.788 & 0.666 & 0.746 & 0.741 & 0.766 & 0.728 & 0.844 & 0.822 & 0.804 & 0.827 & 0.846 & \tbb{0.888} & 0.880 & \trb{0.926} \\
		& mean $F_{\beta}\uparrow$     & 0.513 & 0.576 & 0.542 & 0.523 & 0.617 & 0.604 & 0.713 & 0.756 & 0.735 & 0.765 & 0.790 & 0.824 & \tbb{0.873} & 0.851 & \trb{0.910} \\
		& adp $F_{\beta}\uparrow$     & 0.717 & 0.796 & 0.702 & 0.744 & 0.726 & 0.753 & 0.730 & 0.778 & 0.762 & 0.782 & 0.795 & 0.829 & \tbb{0.866} & 0.863 & \trb{0.915} \\
		& max $E_{\xi}\uparrow$       & 0.850 & 0.890 & 0.773 & 0.851 & 0.856 & 0.870 & 0.881 & 0.932 & 0.928 & 0.893 & 0.910 & 0.923 & \tbb{0.943} & 0.935 & \trb{0.971} \\
		& mean $E_{\xi}\uparrow$     & 0.612 & 0.649 & 0.632 & 0.621 & 0.707 & 0.684 & 0.810 & 0.826 & 0.825 & 0.838 & 0.863 & 0.889 & \tbb{0.933} & 0.902 & \trb{0.948} \\
		& adp $E_{\xi}\uparrow$     & 0.855 & 0.911 & 0.849 & 0.869 & 0.852 & 0.877 & 0.874 & 0.911 & 0.904 & 0.912 & 0.919 & 0.927 & 0.944 & \tbb{0.946} & \trb{0.975} \\
		& $\mathcal{M}\downarrow$ & 0.169 & 0.208 & 0.111 & 0.122 & 0.090 & 0.093 & 0.068 & 0.055 & 0.065 & 0.049 & 0.046 & 0.038 & \tbb{0.030} & 0.033 & \trb{0.021} \\
		\midrule
		\multirow{8}{*}{\begin{sideways}\NLPR\end{sideways}}
		& $S_{\alpha}\uparrow$    & 0.673 & 0.762 & 0.724 & 0.805 & 0.756 & 0.802 & 0.799 & 0.860 & 0.856 & 0.874 & 0.886 & 0.888 & 0.899 & \tbb{0.905} & \trb{0.925} \\
		& max $F_{\beta}\uparrow$     & 0.607 & 0.745 & 0.648 & 0.793 & 0.713 & 0.778 & 0.771 & 0.825 & 0.815 & 0.841 & 0.863 & 0.867 & 0.879 & \tbb{0.885} & \trb{0.914} \\
		& mean $F_{\beta}\uparrow$     & 0.429 & 0.626 & 0.543 & 0.649 & 0.624 & 0.664 & 0.755 & 0.740 & 0.737 & 0.802 & 0.819 & 0.840 & \tbb{0.864} & 0.852 & \trb{0.894} \\
		& adp $F_{\beta}\uparrow$     & 0.535 & 0.736 & 0.614 & 0.665 & 0.692 & 0.744 & 0.747 & 0.724 & 0.730 & 0.795 & 0.796 & 0.823 & \tbb{0.854} & 0.832 & \trb{0.881} \\
		& max $E_{\xi}\uparrow$       & 0.780 & 0.855 & 0.793 & 0.885 & 0.847 & 0.880 & 0.879 & 0.929 & 0.913 & 0.925 & 0.941 & 0.932 & \tbb{0.947} & 0.945 & \trb{0.961} \\
		& mean $E_{\xi}\uparrow$     & 0.578 & 0.719 & 0.684 & 0.745 & 0.742 & 0.755 & 0.851 & 0.840 & 0.841 & 0.887 & 0.902 & 0.918 & \tbb{0.940} & 0.923 & \trb{0.948} \\
		& adp $E_{\xi}\uparrow$     & 0.742 & 0.855 & 0.786 & 0.812 & 0.839 & 0.868 & 0.884 & 0.869 & 0.872 & 0.916 & 0.916 & 0.924 & \tbb{0.941} & 0.931 & \trb{0.956} \\
		& $\mathcal{M}\downarrow$ & 0.179 & 0.081 & 0.117 & 0.095 & 0.091 & 0.085 & 0.058 & 0.056 & 0.059 & 0.044 & 0.041 & 0.036 & \tbb{0.031} & 0.033 & \trb{0.024} \\
		\midrule
		\multirow{8}{*}{\begin{sideways}\SSD\end{sideways}}
		& $S_{\alpha}\uparrow$    & 0.675 & 0.621 & 0.704 & 0.673 & 0.675 & 0.747 & 0.714 & 0.776 & 0.813 & 0.841 & 0.839 & 0.807 & 0.857 & \tbb{0.865} & \trb{0.867} \\
		& max $F_{\beta}\uparrow$     & 0.682 & 0.619 & 0.711 & 0.703 & 0.710 & 0.735 & 0.687 & 0.729 & 0.781 & 0.807 & 0.810 & 0.766 & 0.844 & \tbb{0.846} & \trb{0.849} \\
		& mean $F_{\beta}\uparrow$     & 0.469 & 0.489 & 0.572 & 0.470 & 0.564 & 0.624 & 0.672 & 0.689 & 0.721 & 0.777 & 0.773 & 0.747 & \tbb{0.828} & 0.815 & \trb{0.832} \\
		& adp $F_{\beta}\uparrow$     & 0.656 & 0.613 & 0.679 & 0.674 & 0.693 & 0.724 & 0.694 & 0.710 & 0.748 & 0.791 & 0.767 & 0.726 & \trb{0.821} & 0.790 & \trb{0.821} \\
		& max $E_{\xi}\uparrow$       & 0.785 & 0.736 & 0.786 & 0.779 & 0.800 & 0.828 & 0.807 & 0.865 & 0.882 & 0.894 & 0.897 & 0.852 & 0.906 & \tbb{0.907} & \trb{0.916} \\
		& mean $E_{\xi}\uparrow$     & 0.566 & 0.574 & 0.646 & 0.576 & 0.631 & 0.690 & 0.762 & 0.796 & 0.796 & 0.856 & 0.861 & 0.839 & \trb{0.897} & 0.886 & \tbb{0.896} \\
		& adp $E_{\xi}\uparrow$     & 0.765 & 0.729 & 0.786 & 0.772 & 0.778 & 0.812 & 0.803 & 0.838 & 0.860 & 0.886 & 0.879 & 0.832 & \tbb{0.892} & 0.885 & \trb{0.902} \\
		& $\mathcal{M}\downarrow$ & 0.203 & 0.278 & 0.169 & 0.192 & 0.165 & 0.142 & 0.118 & 0.099 & 0.082 & 0.062 & 0.063 & 0.082 & \tbb{0.058} & 0.059 & \trb{0.050} \\
		\midrule
		\multirow{8}{*}{\begin{sideways}\SIP\end{sideways}}
		& $S_{\alpha}\uparrow$    & 0.732 & 0.727 & 0.683 & 0.717 & 0.628 & 0.653 & 0.720 & 0.716 & 0.833 & 0.842 & 0.835 & 0.850 & 0.806 & \tbb{0.864} & \trb{0.883} \\
		& max $F_{\beta}\uparrow$     & 0.763 & 0.751 & 0.618 & 0.698 & 0.661 & 0.657 & 0.712 & 0.694 & 0.818 & 0.838 & 0.830 & 0.851 & 0.821 & \tbb{0.861} & \trb{0.890} \\
		& mean $F_{\beta}\uparrow$     & 0.542 & 0.571 & 0.499 & 0.568 & 0.515 & 0.464 & 0.702 & 0.608 & 0.771 & 0.814 & 0.803 & 0.821 & 0.811 & \tbb{0.830} & \trb{0.873} \\
		& adp $F_{\beta}\uparrow$     & 0.727 & 0.733 & 0.645 & 0.694 & 0.662 & 0.673 & 0.705 & 0.684 & 0.795 & 0.825 & 0.809 & 0.819 & 0.819 & \tbb{0.829} & \trb{0.875} \\
		& max $E_{\xi}\uparrow$       & 0.838 & 0.853 & 0.743 & 0.798 & 0.771 & 0.759 & 0.819 & 0.829 & 0.897 & 0.901 & 0.895 & 0.903 & 0.875 & \tbb{0.910} & \trb{0.925} \\
		& mean $E_{\xi}\uparrow$     & 0.614 & 0.651 & 0.598 & 0.645 & 0.592 & 0.565 & 0.793 & 0.705 & 0.845 & 0.878 & 0.870 & \tbb{0.893} & 0.844 & \tbb{0.893} & \trb{0.913} \\
		& adp $E_{\xi}\uparrow$     & 0.827 & 0.841 & 0.786 & 0.805 & 0.756 & 0.794 & 0.815 & 0.824 & 0.886 & 0.899 & 0.893 & 0.899 & 0.863 & \tbb{0.901} & \trb{0.920} \\
		& $\mathcal{M}\downarrow$ & 0.172 & 0.200 & 0.186 & 0.167 & 0.164 & 0.185 & 0.118 & 0.139 & 0.086 & 0.071 & 0.075 & 0.064 & 0.085 & \tbb{0.063} & \trb{0.052} \\
		\bottomrule
		\hline
	\end{tabular}
	\label{tab:Results}
\end{table*}

\begin{figure*}[t!]
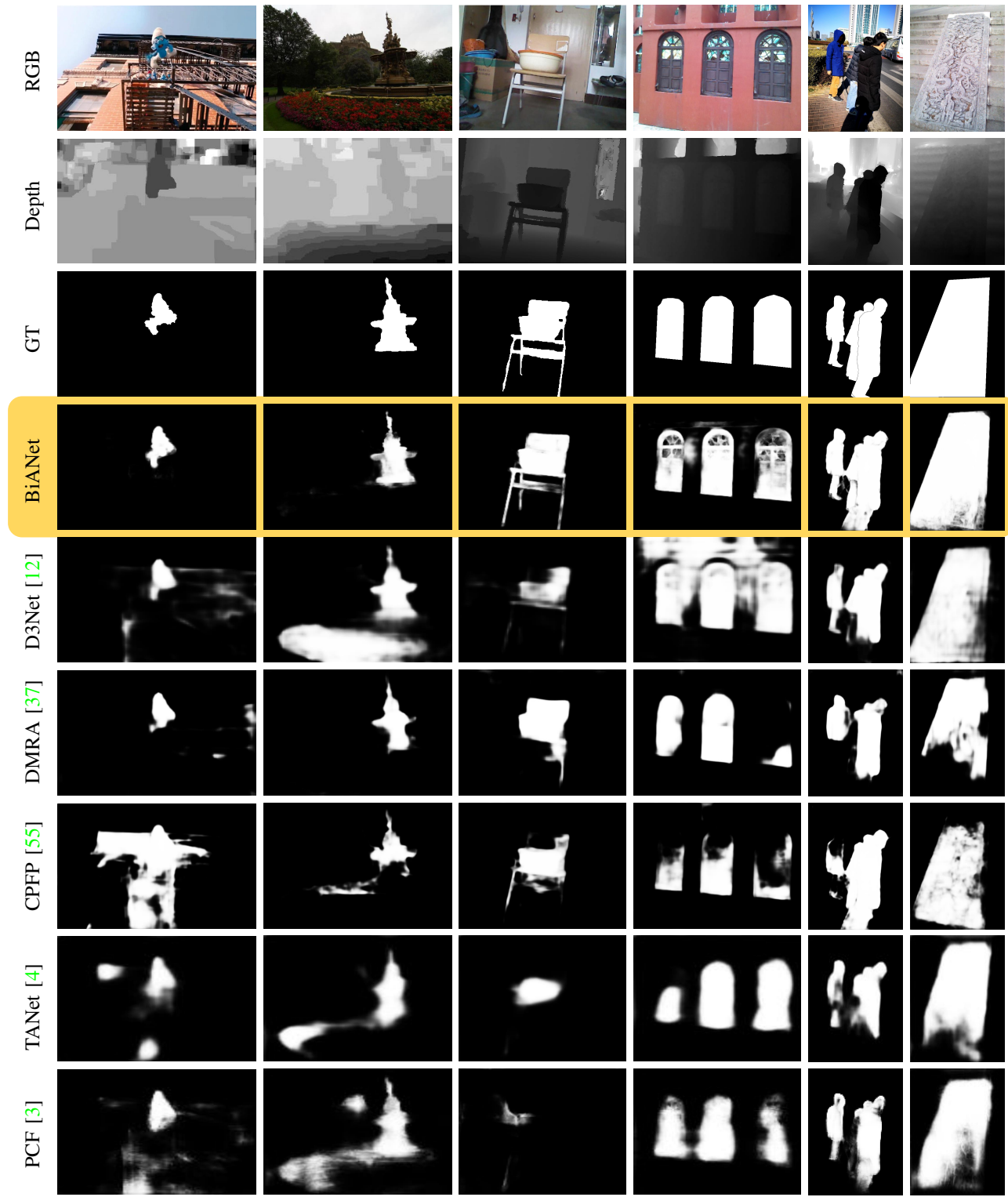

	\centering
	\begin{overpic}[width=.95\linewidth]{Pics/SalMaps.pdf}
	\put(1.8,91.5){\rotatebox{90}{RGB}}
	\put(1.8,80){\rotatebox{90}{Depth}}
	\put(1.8,70){\rotatebox{90}{GT}}
	\put(1.8,57.6){\rotatebox{90}{BiANet}}
	\put(1.8,45.5){\rotatebox{90}{\DTNet}}
	\put(1.8,34.5){\rotatebox{90}{\DMRA}}
	\put(1.8,24){\rotatebox{90}{\CPFP}}
	\put(1.8,12.5){\rotatebox{90}{\TANet}}
	\put(1.8,2.8){\rotatebox{90}{\PCF}}
	\end{overpic}
	\caption{
	\textbf{Visual comparison of BiANet with other top 5 methods}. The inputs include difficult scenes of tiny objects (column 1), complex background (column 1 and 2), complex texture (column 3), low contrast (column 2 and 6), low-quality or confusing depth (column 2, 4, and 6), and multiple objects (column 4 and 5).}
	\vspace{-2pt}
\label{fig:salmaps}
\end{figure*}

\section{Experiments}
\label{sec:exp}
\noindent
\subsection{Evaluation Protocols}

\subsubsection{Evaluation datasets} 
We conduct experiments on six widely used RGB-D based SOD datasets. 
\NJU~and \NLPR~are two popular large-scale RGB-D SOD datasets containing 1985 and 1000 images, respectively. 
\DES~contains 135 indoor images with fine structures collected with Microsoft Kinect~\cite{zhang2012microsoft}.
\STERE~contains 1000 internet images, and the corresponding depth maps are generated by stereo images using a sift flow algorithm~\cite{liu2011sift}.
\SSD~is a small-scale but high-resolution dataset with 400 images in $960\times1080$ resolution.
\SIP~is a high-quality RGB-D SOD dataset with 929 person images.

\subsubsection{Evaluation metrics}
We employ 9 metrics to comprehensively evaluate these methods.
\textbf{Precision-Recall (PR) curve}~\cite{powers2011evaluation} shows the precision and recall performances of the predicted saliency map at different binary thresholds.
\textbf{F-measure}~\cite{achanta2009frequency} is computed by the weighted harmonic mean of the thresholded precision and recall. We employ maximum F-measure (max $F_{\beta}$), mean F-measure (mean $F_{\beta}$), and adaptive F-measure (adp $F_{\beta}$).
\textbf{Mean Absolute Error} (MAE, $\mathcal{M}$)~\cite{perazzi2012} directly estimates the average pixel-wise absolute difference between the prediction and the binary ground-truth map.
\textbf{S-measure ($S_{\alpha}$)}~\cite{fan2017structure} is an advanced metric,
which takes the region-aware and object-aware structural similarity into consideration.
\textbf{E-measure}~\cite{Fan2018Enhanced} is the recent proposed Enhanced alignment measure in the binary map evaluation
field, which combines local pixel values with the image level
mean value in one term, jointly capturing image-level
statistics and local pixel matching information.
Similar to $F_{\beta}$, 
we employ the maximum E-measure (max $E_{\xi}$), mean E-measure (mean $E_{\xi}$), and adaptive E-measure (adp $E_{\xi}$).


\begin{figure*}[t]
	\centering
	\begin{overpic}[width=0.96\textwidth]{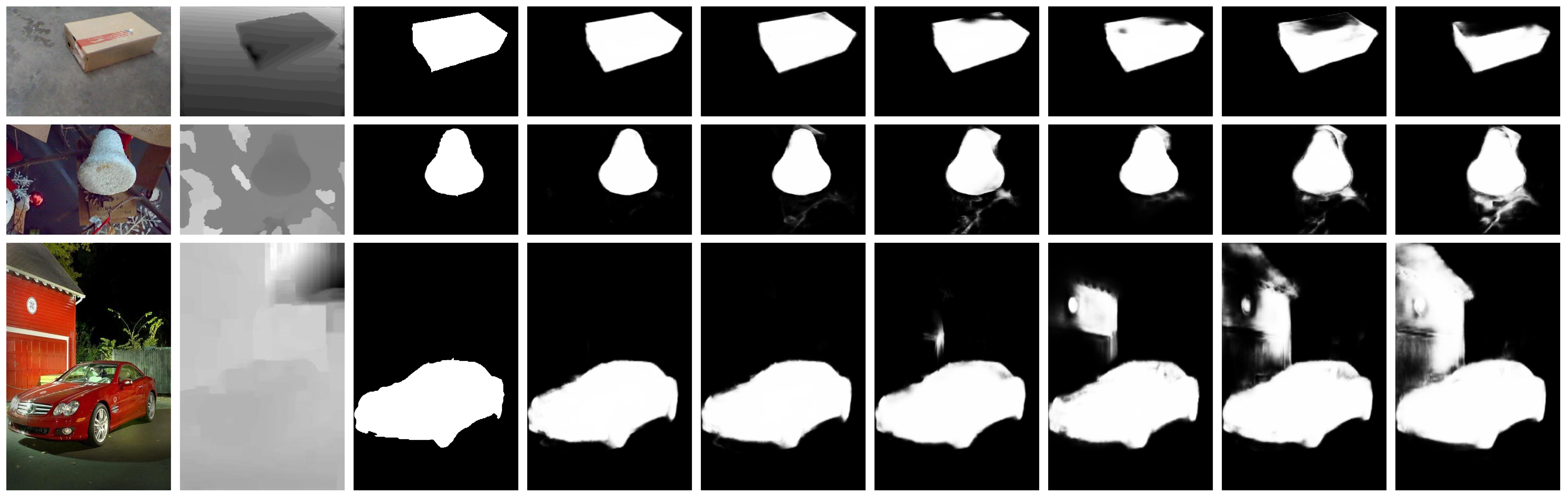}
		\put(4,32){\small{RGB}}
		\put(15,32){\small{Depth}}
		\put(27,32){\small{GT}}
		\put(37,32){\small{No.\ 6}}
		\put(48,32){\small{No.\ 5}}
		\put(59,32){\small{No.\ 4}}
		\put(70,32){\small{No.\ 3}}
		\put(81.5,32){\small{No.\ 2}}
		\put(92,32){\small{No.\ 1}}
	\end{overpic}
	\caption{
		\textbf{Visual comparison in the ablation studies}.\ The candidate mechanisms are deep information (Dep), foreground-first attention (FF), background-first attention (BF), and multi-scale extension (ME).
		No.\ 6: (Dep + FF + BF + ME).
		No.\ 5: (Dep + FF + BF).
		No.\ 4: (Dep + BF).
		No.\ 3: (Dep + FF).
		No.\ 2: Dep.
		No.\ 1: Baseline.
	}
	\label{fig:ABL}
\end{figure*}

\begin{table}[t]
	\centering
	\small
	\caption{
		\textbf{Ablation analysis for the proposed architecture} on the \textit{NJU2K} and \textit{STERE} datasets.
		The candidate mechanisms are deep information (Dep), foreground-first attention (FF), background-first attention (BF), and multi-scale extension (ME). 
		ME is applied on the top three level features.
	}
	\renewcommand{\arraystretch}{1.0}
	\renewcommand{\tabcolsep}{1.82mm}
	\begin{tabular}{c|cccc|cccc}
		\hline\toprule
		\multirow{2}{*}{\#} & \multicolumn{4}{c|}{Candidates}  & \multicolumn{2}{c}{\NJU} & \multicolumn{2}{c}{\STERE}  \\
		& Dep            & FF        & BF          & ME         & $F_{\beta}\uparrow$ & $S_{\alpha}\uparrow$  & $F_{\beta}\uparrow$ & $S_{\alpha}\uparrow$ \\
		\hline
		No.\ 1  &            &             &            &       &   0.881     &   0.885   &  0.882  &  0.893        \\
		No.\ 2  &\checkmark  &             &            &       &   0.903     &   0.904   &  0.887  &  0.894   \\
		No.\ 3  &\checkmark  &  \checkmark &            &       &   0.908     &   0.908   &  0.895  &  0.901    \\
		No.\ 4  &\checkmark  &             & \checkmark &       &   0.910     &   0.908   &  0.892  &  0.900    \\
		No.\ 5  &\checkmark  &  \checkmark & \checkmark &       &   0.915     &   0.913   &  0.897  &  0.903    \\
		\hline
		No.\ 6  &\checkmark  &  \checkmark & \checkmark &  \checkmark  &  \textbf{0.920}  &   \textbf{0.915} &  \textbf{0.898}  &   \textbf{0.904} \\
		\bottomrule
		\hline
	\end{tabular}
	\label{tab:ABL}
\end{table}

\begin{table}[t]
	\centering
	\small
	\caption{\textbf{Improvements of accuracy by our BAM in each side outputs} compared with No.\ 2 (without BAM \& MBA).  
	}
	\renewcommand{\arraystretch}{1.0}
	\renewcommand{\tabcolsep}{1.65mm}
	\begin{tabular}{c|ccccc|c}
		\hline\toprule
		BAM & Level-1   & Level-2    & Level-3  & Level-4   &  Level-5 & No.\ 2     \\
		\hline
		$S_{\alpha}\uparrow$      & 0.908 & 0.909 &  0.908 & 0.906  & 0.904 & 0.904 \\
		$F_{\beta}\uparrow$       & 0.910 & 0.911 &  0.909 & 0.905  & 0.904 & 0.903 \\
		$E_{\xi}\uparrow$         & 0.944 & 0.945 &  0.943 & 0.943  & 0.941 & 0.942 \\
		$\mathcal{M}\downarrow$   & 0.043 & 0.043 &  0.044 & 0.044  & 0.045 & 0.046 \\
		\bottomrule 
		\hline
	\end{tabular}
	\label{tab:BAM_level}
\end{table}

\begin{table}[t!]
	\centering
	\small
	\caption{\textbf{Improvements of accuracy by our MBAM in each side outputs} compared with No.\ 2 (without BAM \& MBAM).  
	}
	\renewcommand{\arraystretch}{1.0}
	\renewcommand{\tabcolsep}{1.5mm}
	\begin{tabular}{c|ccccc|c}
		\hline\toprule
		MBAM    & Level-1   & Level-2    & Level-3  & Level-4   &  Level-5 & No.\ 2     \\
		\hline
		$S_{\alpha}\uparrow$      & 0.908 & 0.909 & 0.910  & 0.910 & 0.910 & 0.904 \\
		$F_{\beta}\uparrow$       & 0.909 & 0.912 & 0.909  & 0.911 & 0.911 & 0.903 \\
		$E_{\xi}\uparrow$         & 0.944 & 0.945 & 0.945  & 0.946 & 0.947 & 0.942 \\
		$\mathcal{M}\downarrow$   & 0.044 & 0.043 & 0.042  & 0.042 & 0.042 & 0.046 \\
		\bottomrule 
		\hline
	\end{tabular}
	\label{tab:MBAM_level}
\end{table}

\begin{table*}[t]
	\centering
	\small
	\renewcommand{\arraystretch}{1.2}
	\renewcommand{\tabcolsep}{4.5mm}
	\caption{\textbf{Accuracy and calculation cost analysis for MBAM}.\ $\times 0 \sim \times 5$ means the number of MBAMs, which are applied from high levels to low levels.
		FPS denotes Frames Per Second.
		Params means the size of parameters.
		FLOPs = Floating Point Operations.
		The accuracy metrics $F_{\beta}$ and $\mathcal{M}$ are evaluated on the \textit{NJU2K} dataset.
		The calculation cost metrics FPS and FLOPs are tested at $224 \times 224$ resolution.
		Note that, $\times$3 is  the default setting in \secref{sec:SOTA}.}
	\begin{tabular}{r|cccccc|cc}
		\hline\toprule
		& $\times 0$ & $\times 1$  & $\times 2$ & $\times 3$  &  $\times 4$  &  $\times 5$ &\DTNet  &\DMRA \\
		\hline
		$F_{\beta}\uparrow$      & 0.914      &  0.917      &  0.918     & 0.920       &  0.920  &  0.921     & 0.887   & 0.886\\
		$\mathcal{M}\downarrow$  & 0.041      &  0.040      &  0.040     & 0.039       &  0.038  &  0.039     & 0.051   & 0.051\\
		FPS$\uparrow$            & $\sim$80   &  $\sim$65   & $\sim$55   & $\sim$50    & $\sim$42&  $\sim$34  &$\sim$55 & $\sim$40 \\
		Params $\downarrow$      & 45.0M      &  46.9M      &  48.7M     & 49.6M       &  50.1M  &  50.4M     & 145.9M  &59.7M\\
		FLOPs  $\downarrow$      & 34.4G      &  35.0G      &  36.2G     & 39.1G       &  45.2G  &  58.4G     & 55.7G   &121.0G\\
		\bottomrule
		\hline
	\end{tabular}
	\label{tab:MBAM}
\end{table*}

\begin{table}[t]
	\centering
	\small
	\renewcommand{\arraystretch}{1.25}
	\renewcommand{\tabcolsep}{0.7mm}
    \caption{Performances of our BiANet based on different backbones. VGG-11 and VGG-16 is the VGG network proposed in \cite{simonyan2015vgg}.
    ResNet-50 is proposed in \cite{he2016deep}.
    Res2Net-50 is proposed in \cite{gao2019res2net}.
	}
	\begin{tabular}{cc|cccc}
		\hline\toprule
		\multicolumn{2}{c|}{Backbone} & ~VGG-11~ & ~~VGG-16~~ & ResNet-50 & Res2Net-50 \\
		\hline
		\multicolumn{2}{c|}{FPS}  & 60 & 50 & 25 & 23   \\
		\hline
		\multirow{4}{*}{\begin{sideways}\NJU\end{sideways}}
		& $S_{\alpha}\uparrow$    & 0.912 & 0.915 & 0.917 & 0.923   \\
		& $F_{\beta}\uparrow$     & 0.913 & 0.920 & 0.920 & 0.925   \\
		& $E_{\xi}\uparrow$       & 0.947 & 0.948 & 0.949 & 0.952   \\
		& $\mathcal{M}\downarrow$ & 0.040 & 0.039 & 0.036 & 0.034   \\
		\hline
		\multirow{4}{*}{\begin{sideways}\STERE\end{sideways}}
		& $S_{\alpha}\uparrow$    & 0.899 & 0.904 & 0.905 & 0.908   \\
		& $F_{\beta}\uparrow$     & 0.892 & 0.898 & 0.899 & 0.904   \\
		& $E_{\xi}\uparrow$       & 0.941 & 0.942 & 0.943 & 0.942   \\
		& $\mathcal{M}\downarrow$ & 0.045 & 0.043 & 0.040 & 0.039   \\
		\hline
		\multirow{4}{*}{\begin{sideways}\DES\end{sideways}}
		& $S_{\alpha}\uparrow$    & 0.943 & 0.931 & 0.930 & 0.942  \\
		& $F_{\beta}\uparrow$     & 0.938 & 0.926 & 0.927 & 0.942  \\
		& $E_{\xi}\uparrow$       & 0.979 & 0.971 & 0.968 & 0.978  \\
		& $\mathcal{M}\downarrow$ & 0.019 & 0.021 & 0.021 & 0.017  \\
		\hline
		\multirow{4}{*}{\begin{sideways}\NLPR\end{sideways}}
		& $S_{\alpha}\uparrow$    & 0.927 & 0.925 & 0.926 & 0.929  \\
		& $F_{\beta}\uparrow$     & 0.914 & 0.914 & 0.917 & 0.919  \\
		& $E_{\xi}\uparrow$       & 0.951 & 0.961 & 0.962 & 0.963  \\
		& $\mathcal{M}\downarrow$ & 0.024 & 0.024 & 0.023 & 0.023  \\
		\hline
		\multirow{4}{*}{\begin{sideways}\SSD\end{sideways}}
		& $S_{\alpha}\uparrow$    & 0.861 & 0.867 & 0.863 & 0.863  \\
		& $F_{\beta}\uparrow$     & 0.839 & 0.849 & 0.843 & 0.843  \\
		& $E_{\xi}\uparrow$       & 0.899 & 0.916 & 0.911 & 0.901  \\
		& $\mathcal{M}\downarrow$ & 0.054 & 0.050 & 0.048 & 0.050  \\
		\hline
		\multirow{4}{*}{\begin{sideways}\SIP\end{sideways}}
		& $S_{\alpha}\uparrow$    & 0.877 & 0.883 & 0.887 & 0.889  \\
		& $F_{\beta}\uparrow$     & 0.882 & 0.890 & 0.890 & 0.893  \\
		& $E_{\xi}\uparrow$       & 0.924 & 0.925 & 0.926 & 0.928  \\
		& $\mathcal{M}\downarrow$ & 0.054 & 0.052 & 0.047 & 0.047  \\
		\bottomrule
		\hline
	\end{tabular}
	\label{tab:backbone}
\end{table}

\subsection{Comparison with State-of-the-Arts}
\label{sec:SOTA}
\subsubsection{Comparison methods}
We compared with 14 state-of-the-art RGB-D SOD methods, including 5 traditional methods: 
\ACSD, LBE~\cite{feng2016local}, DCMC~\cite{cong2016saliency}, MDSF~\cite{song2017depth}, and SE~\cite{guo2016salient},
and 9 DNN-based methods: DF~\cite{qu2017rgbd}, AFNet~\cite{wang2019adaptive}, CTMF~\cite{han2017cnns}, MMCI~\cite{chen2019multi}, PCF~\cite{chen2018progressively}, TANet~\cite{chen2019three}, CPFP~\cite{zhao2019Contrast},  DMRA~\cite{piao2019depth}, and D3Net~\cite{fan2019D3Net}.
The codes and saliency maps of these methods are provided by the authors.

\subsubsection{Quantitative evaluation}
The complete quantitative evaluation results are listed in \tabref{tab:Results}.
The comparison methods are presented from right to left according to the comprehensive performance of these metrics,
where the lower the value of MAE ($\mathcal{M}$), the better the effect of the model.
The other metrics are the opposite.
We also plot the PR curves of these methods in \figref{fig:PR}.
One can see that our BiANet achieves remarkable advantages over the comparison methods.
\DMRA~and \DTNet~are well-matched in these datasets.
On the large-scaled \NJU~and \NLPR~datasets, our BiANet outperforms the second best with $\sim$3\% improvement on max $F_{\beta}$.
On the \DES~dataset, Compared to methods which are heavily dependent on depth information,
our proposed BiANet also has a 3.8\% improvement on max $F_{\beta}$.
This indicates that our BiANet can make more efficient use of depth information.
Although the \SSD~dataset is high-resolution, the quality of the depth map is poor.
Our BiANet still exceeds \DTNet, which is specifically designed for robustness to low-quality depth maps.
Our BiANet also performs the best on the \SIP, which is a challenging dataset with complex scenes and multiple objects.

\subsubsection{Qualitative results}
To further demonstrate the effectiveness of our BiANet, we visualized the saliency maps of our BiANet and other top 5 methods in~\figref{fig:salmaps}.
One can see that the target object in the 1st column is tiny, and its white shoes and hat are hard to distinguish from the background. 
Our BiANet effectively utilizes the depth information, 
while the others are disturbed by RGB background clutter.
The inputs in the 2nd column are challenging because the depth map is mislabeled, and the RGB image was taken in a dark environment with low contrast.
Our BiANet successfully detects the target sculpture and eliminates the interference of flowers and the base of the sculpture,
while D3Net mistakenly detects a closer rosette, and DMRA loses the part of the object that is similar to the background.
The 3rd column shows the ability of our BiANet to detect complex structures of salient objects.
Among these methods, only our BiANet completely discover the chairs, including the fine legs.
The 4th column is a multi-object scene.
Because there are no depth differences between the three salient windows below and the wall, 
they are not reflected on the depth map, but the three windows above are clearly observed on the depth map.
In this case, the depth map will mislead subsequent segmentation.
Our BiANet detects multiple objects from RGB images with less noise.
The 5th column is also a multi-object scene. 
The bottom half of depth map is confused with the interference from the ground.
Thus, detecting the legs of these persons in the image is very difficult.
However, our BiANet successfully detected all the legs.
The last row is a large-scale object whose color and depth map are not distinguished.
Large scale, low color contrast and lack of discriminative depth information make the scene very challenging.
Fortunately, our BiANet is robust on this scene.

\subsection{Ablation Study}
\label{sec:expvalid}
\vspace{-1mm}
In this section, we mainly investigate: 1) the benefits of bilateral attention mechanism to our BiANet; 2) the effectiveness of BAM in different levels to our BiANet for RGB-D SOD; 3) the further improvements of MBAM in different levels to our BiANet; 4) the benefits of combining BAM and MBAM for RGB-D SOD; and 5) the impact of different backbones to our BiANet for RGB-D SOD.\

\subsubsection{Effectiveness of bilateral attention}
We conduct ablation studies on the large-scaled \textit{NJU2K} and \textit{STERE} datasets to investigate the contributions of different mechanisms in the proposed method.
The baseline model used here contains a VGG-16 backbones and a residual refine structure.
It takes RGB images as input without depth information.
The performance of our basic network without any additional mechanisms is illustrated in \tabref{tab:ABL} No.\ 1.
Based on the network, we gradually add different mechanisms and test various combinations.
These candidates are depth information (Dep), foreground-first attention (FF), background-first attention (BF), and multi-scale extension (ME).
In \tabref{tab:ABL} No. 3,
by applying FF, the performance is improved to some extent, 
It benefits from the foreground cues being learned effectively by shifting the attention to the foreground objects.
This is also reflected in \figref{fig:ABL}.
The foreground objects are detected more accurately;
however, without good understanding on background cues,
it tend to mistake some background objects, such as the red house in the third row, 
or cannot find complete foreground objects as lack of exploration on background regions.
We get a similar accuracy when using the BF only, as shown in No.\ 4.
It excels at distinguishing between salient areas and non-salient areas in the background, 
and can help to find more complete regions of the salient object in the uncertain background;
however, 
too much attention focusing on the background and without a good understanding of the foreground cues, 
it leads that sometimes background noise is introduced.
When we combine FF together with BF to form our BAM and apply it in all side outputs, the performance boosts.
We can see that BAM increases S-measure by 0.9\% and max F-measure by 1.2\% compared with No.\ 2.
When we replace the top three levels BAMs with MBAMs, 
the performance further improved.
In \figref{fig:ABL}, 
compared to the performance of No.\ 2 without BAM, the detected salient objects of No.\ 6 possess higher confidence, sharper edges, and less background noise.

\subsubsection{Effectiveness of BAM with different levels}
In order to verify that our BAM module is effective at each feature level,
we apply BAM to each side output of the No.\ 2 model's feature extractor, respectively. 
That is, in each experiment, BAM is applied to one side output, while the others undergo general convolutions.
From~\tabref{tab:BAM_level},
we can see that the BAMs in every layer facilitate a universal improvement on detection performance.
In addition, we find that BAM applied in the lower levels contributes more to the results.

\subsubsection{Effectiveness of MBAM in different levels}
\label{sec:abl_mbam}
In \tabref{tab:ABL},
compared with No. 5, No. 6 carry out multi-scaled extension on its higher three levels $\{\mathbf{F}_3,\mathbf{F}_4,\mathbf{F}_5\}$.
This extension effectively improves the performance of the model.
In order to better show the gain of MBAM in each level features, similar to \tabref{tab:BAM_level}, 
we apply MBAM to each side output of the No. 2 model, respectively.
The experimental results are recorded in \tabref{tab:MBAM_level}, where
different levels of MBAM bring different degrees of improvement to the results.
Comparing \tabref{tab:BAM_level} and \tabref{tab:MBAM_level}, 
we can see a more interesting phenomenon that 
the BAM applied in the lower level brings more improvement 
while the MBAM applied in the higher level is more effective.

\subsubsection{Cooperation between BAM and MBAM}
The observation above guides us that when using BAM and MBAM in cooperation, 
we should give priority to multi-scale expansion of higher-level BAM.
Therefore, we expand BAM from top to bottom until all BAMs are converted into MBAMs.
We record the final detection performance and calculation cost during the gradual expansion in \tabref{tab:MBAM}.
%
We start from the highest level,
and gradually increase the number of MBAMs to three.
We can see that the effect on the model is a steady improvement, but the computing cost is also increased.
At the lower levels, adding MBAM has no obvious effect.
This phenomenon is in line with our expectation.
Besides, due to the high resolution, 
the extension of lower-level BAM will increase the calculation cost and reduce the robustness.
The selection of the number of MBAM needs to balance the accuracy and speed requirements of the application scenario.
In scenarios with higher speed requirements, we recommend not to use MBAM. 
Our most lightweight model can achieve $\sim$80fps while ensuring significant performance advantages.
The parameter size and FLOPs are superior to the SOTA methods \DTNet~and \DMRA.
In scenarios where high accuracy is required, we suggest applying less than three MBAMs on higher-level features.

\subsubsection{Performances under different backbones}
\label{sec:abl_combination}
We implement the BiANet based on some other widely-used backbones 
to demonstrate the effectiveness of the proposed bilateral attention mechanism on different feature extractors.
Specifically,
in addition to VGG-16~\cite{simonyan2015vgg}, 
we provide the results of BiANet on VGG-11~\cite{simonyan2015vgg}, ResNet-50~\cite{he2016deep}, and Res2Net-50~\cite{gao2019res2net}.
Compared with VGG-16, VGG-11 is a lighter backbone.
As shown in \tabref{tab:backbone}, 
although the accuracy is slightly lower than VGG-16, 
it still reaches SOTA with a faster speed.
BiANet with stronger backbones will bring more remarkable improvements. 
For example, 
when we employ ResNet-50 like \DTNet\ as backbone, 
our BiANet brings 1.5\% improvement on \NJU\ in terms of the MAE compared with the \DTNet.
When armed with Res2Net-50~\cite{gao2019res2net}, 
BiANet achieves 3.8\% improvement on \NJU\ in terms of the max F-measure compared with the SOTA methods.

\begin{figure}[t]
	\centering
	\begin{overpic}[width=.95\columnwidth]{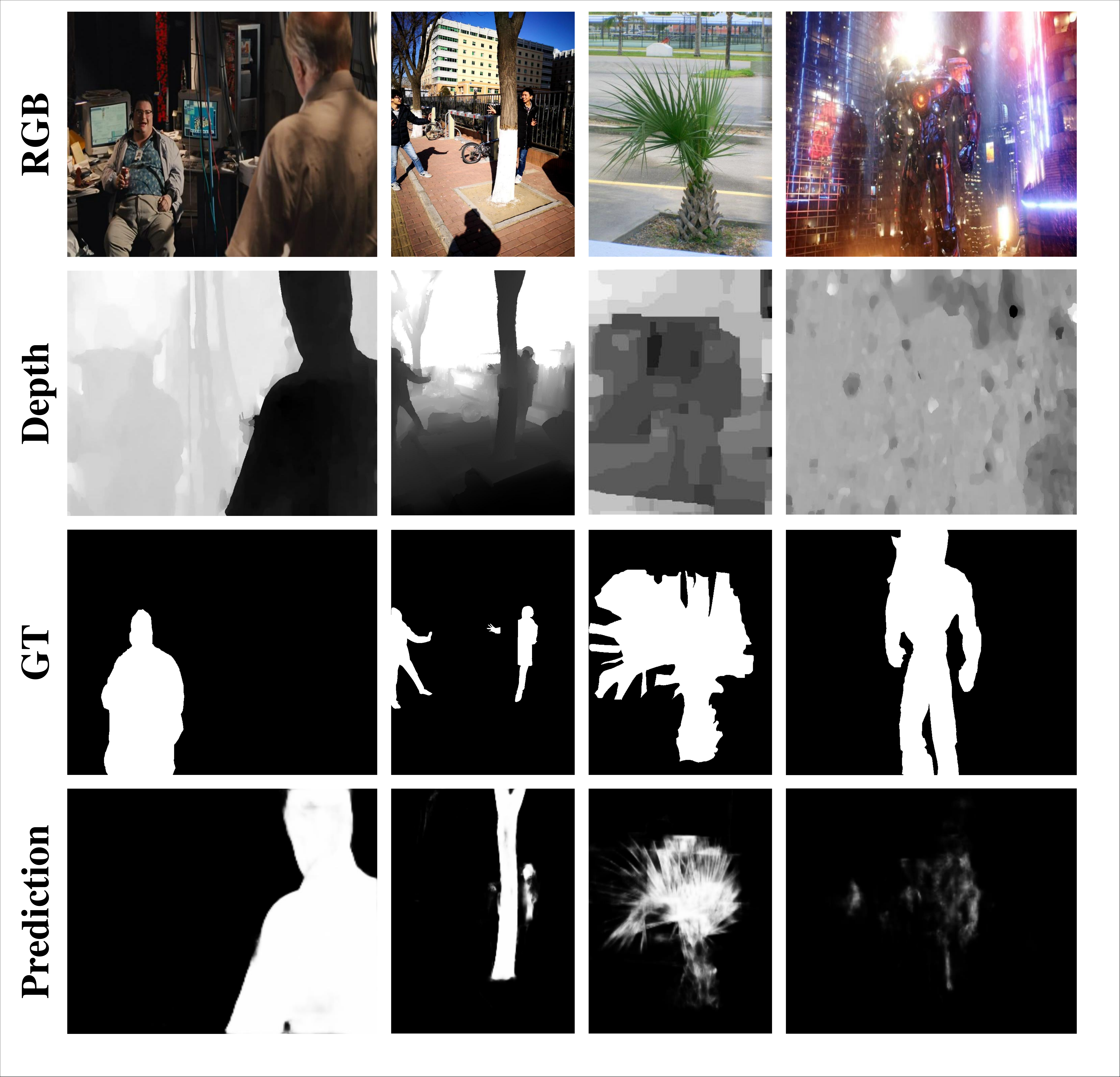} 
	\end{overpic}
	\caption{
		\textbf{Failure cases of BiANet in extreme environments}. 
		In the first two columns, as the objects closer to the observer are not the targets, the depth maps provide misleading information.
		In the last two columns, the BiANet fails lead by the confusing RGB information and coarse depth maps.
	}
	\label{fig:Failures}
\end{figure}

\subsection{Failure Case Analysis}
In \figref{fig:Failures}, we illustrate some failure cases when our BiANet works in some extreme environments.
BiANet explores the saliency cues bilaterally in the foreground and background regions with the relationship provided by depth information.
However, when the foreground regions indicated by depth information do not belong to the salient object, 
it is likely to mislead the prediction.
The first two columns in \figref{fig:Failures} are typical examples,
where our BiANet mistakenly takes the object close to the observer as the target, and gives the wrong prediction.
The other situation that may cause failure is when BiANet encounters coarse depth maps in complex scenarios ( see the last two columns).
In the third column, the depth map provides inaccurate spatial information, which affects the detection of details.
In the last column, the inaccurate depth map and the confusing RGB information make BiNet fail to locate the target object.

\section{Conclusion}
\label{conclusion}

In this paper, we propose a fast yet effective bilateral attention network (BiANet) for RGB-D saliency object detection (SOD) task.
To better utilize the foreground and background information, we propose a bilateral attention module (BAM) to comprise the dual complementary of foreground-first attention and 
background-first attention mechanisms. 
To fully exploit the multi-scale techniques, we extend our BAM module to its multi-scale version (MBAM), capturing better global information. 
Extensive experiments on six benchmark datasets demonstrated that our BiANet, benefited by our BAM and MBAM modules, outperforms previous state-of-the-art methods on RGB-D SOD, in terms of quantitative and qualitative performance. 
The proposed BiANet runs at real-time speed on a single GPU, making it a potential solution for various real-world applications.
\ifCLASSOPTIONcaptionsoff
  \newpage
\fi



\bibliographystyle{ieee_fullname} 
\bibliography{TIP_REF}


\vfill
\end{document}